\documentclass{article}
\usepackage[utf8]{inputenc}
\usepackage[T1]{fontenc}
\usepackage{geometry}
\usepackage{translator}
\usepackage[catalan, spanish, british]{babel}
\usepackage{johd}

\usepackage{setspace}
\usepackage[colorinlistoftodos]{todonotes}
\usepackage{hyperref}
\usepackage{soul}
\usepackage{underscore}
\usepackage{array, multirow}
\usepackage{tabularx, tabulary, longtable}
\usepackage{tipa}
\usepackage{multicol}
\usepackage{makecell}
\usepackage{booktabs}
\usepackage{enumitem}
\usepackage{ragged2e}
\usepackage{xspace}
\usepackage{etoolbox}

\newcolumntype{Y}{>{\RaggedRight\arraybackslash}X}

\let\oldbibliography\thebibliography
\renewcommand{\thebibliography}[1]{%
  \oldbibliography{#1}%
  \setlength{\itemsep}{8pt}%
  \setlength{\baselineskip}{11.5pt}%
}

\newcommand{\mapu}[1]{\emph{#1}}
\newcommand{\mapuex}[1]{\footnotesize\emph{#1}}
\newcommand{\gloss}[1]{\footnotesize\textsc{\lowercase{#1}}}
\newcommand{\example}[1]{\textbf{E.#1}\xspace}

\usepackage{titlesec}
\titleformat{\section}
  {\normalfont\Large\bfseries}
  {\thesection}{1em}{}
\titleformat{\subsection}
  {\normalfont\large\bfseries}
  {\thesubsection}{1em}{}
\titleformat{\subsubsection}
  {\normalfont\normalsize\bfseries}
  {\thesubsubsection}{1em}{}

\hypersetup{
    colorlinks=true,
    linkcolor=blue,
    citecolor=violet,
    urlcolor=blue,
    pdftitle={Lexical categories of stem-forming roots in Mapudüngun verb forms},
    pdfauthor={Andrés Chandía}
}

\begin{document}

\begin{center}
  {\Huge\bfseries Lexical categories of stem-forming roots in \mapu{Mapudüngun} verb forms\par}
  \vspace{0.5cm}
  {\large\textit{A preliminary study for valency classification}\par}
  {\large
  Andrés Chandía\\[2pt]
  \small\href{mailto:andres@chandia.net}{\texttt{andres@chandia.net}}\\[8pt]
  Department of Catalan Philology and General Linguistics\\ 
  University of Barcelona\\[4pt]
  PhD programme: Cognitive Science and Language\\[12pt]
  {\small Supervised by}\\
  {\small Dr. Elisabet Comelles \href{mailto:elicomelles@ub.edu}{\texttt{(elicomelles@ub.edu)}} and\\
  {\small Dr. Irene Castellón \href{mailto:icastellon@ub.edu}{\texttt{(icastellon@ub.edu)}}}}
  }
  \vfill
\end{center}
\thispagestyle{empty}

\vspace{0.5cm}

\begin{abstract}
\noindent After developing a computational system for morphological analysis of the \mapu{Mapuche} language, and evaluating it with texts from various authors and styles, it became necessary to verify the linguistic assumptions of the source used as the basis for implementing this tool.
\noindent In the present work, the primary focus is on the lexical category classification of \mapu{Mapudüngun} roots recognised as verbal in the source utilised for the development of the morphological analysis system.
\noindent The results of this lexical category revision directly benefit the computational analyser, as they are implemented as soon as they are verified. Additionally, it is hoped that these results will help clarify some uncertainties about lexical categories in the \mapu{Mapuche} language.
\noindent This work addresses a preliminary task to identify the valency of true verbal roots, the results of which will be presented in a subsequent work that complements this article.
\end{abstract}


\selectlanguage{spanish}
\begin{abstract}
\noindent Tras haber desarrollado un sistema computacional para el análisis morfológico de la lengua \mapu{Mapuche} y de evaluarlo con textos de diversos autores y estilos, se ha visto la necesidad de verificar los postulados lingüísticos de la fuente utilizada como base para implementar esta herramienta.
\noindent En el presente trabajo, el enfoque principal es la clasificación de la categoría léxica de las raíces del \mapu{Mapudüngun} reconocidas como verbales en la fuente utilizada para el desarrollo del sistema de análisis morfológico.
\noindent Los resultados de esta revisión categorial benefician directamente al analizador computacional, ya que se implementan tan pronto como se verifican. Además, se espera que estos resultados ayuden a aclarar algunas incertidumbres sobre las categorías léxicas en la lengua \mapu{Mapuche}.
\noindent Este trabajo aborda una tarea preliminar a la identificación de la valencia de las raíces verbales, cuyos resultados se presentarán en un trabajo posterior que complementa este artículo.
\end{abstract}

\noindent\textbf{Palabras clave:} Mapuche; Mapudüngun; FST; transductor de estados finitos; morfología; morfosintaxis; afijos; sufijos; morfemas; verbos; valencia; raíces verbales; categorías gramaticales; categorías léxicas

\selectlanguage{catalan}
\begin{abstract}
\noindent Després d'haver desenvolupat un sistema computacional d'anàlisi morfològica de la llengua \mapu{Mapuche}, i d'avaluar-lo amb textos de diversos autors i estils, es va veure la necessitat de verificar els postulats lingüístics de la font utilitzada com a base per implementar aquesta eina.
\noindent En el present treball, l'enfocament principal és la classificació de la categoria lèxica de les arrels del \mapu{Mapudüngun} reconegudes com a verbals en la font utilitzada per al desenvolupament del sistema d'anàlisi morfològica.
\noindent Els resultats d'aquesta revisió categorial beneficien directament l'analitzador computacional, ja que s'implementen tan bon punt es verifiquen. A més, s'espera que aquests resultats ajudin a aclarir algunes incerteses sobre les categories lèxiques en la llengua \mapu{Mapuche}.
\noindent Aquest treball aborda una tasca preliminar a la identificació de la valència de les arrels verbals, els resultats de la qual es presentaran en un treball posterior que complementa aquest article.
\end{abstract}

\noindent\textbf{Paraules clau:} Maputxe; Mapudüngun; FST; transductor d'estats finits; morfologia; morfosintaxi; afixos; sufixos; morfemes; verbs; valència; arrels verbals; categories gramaticals; categories lèxiques

\selectlanguage{british}
\vspace{1cm}
\tableofcontents
\listoftables

\vspace{1.5cm}

\subsection*{Note on the spelling of \mapu{Mapuche} names}
In line with the \textit{alfabeto Mapuche unificado} (AMU), we write \mapu{Mapuche} names and surnames in their original form (\mapu{Longkon, Koña}) rather than using the standard bibliographical Castilianisations (Loncón, Coña). This choice reflects a commitment to recognise and respect the \mapu{Mapuche} people and their language. The references section includes the standardised academic forms in the \texttt{note} field to facilitate source location.

\vspace{1.5cm}

\section{\label{sec.01} Introduction}

This article details the methods used for the reclassification of the lexical roots involved in \mapu{Mapuche} verb themes and presents the results obtained from this process. This task is essential to accurately determine the valency (in a subsequent work, \citealp{chandia2026}) of roots that are genuinely verbal. Without this step, there is a risk of incorrectly attributing verbal properties to roots that do not belong to this lexical category.

\mapu{Mapudüngun}\footnote{The spelling \textit{mapudüngun} follows the AMU alphabet, where \textit{ü} stands for the central vowel /\textschwa/ (a pronunciation attested from Valdivia (1606) to \citet{navarro1993}, including Febrés and Augusta).\vspace{0.2cm}} is a polysynthetic and predominantly suffixing language. For a working definition of terms such as \textit{synthesis} and \textit{agglutination}, the reader is referred to the companion article on valency (\citealp{chandia2026}: §1, footnotes 4–5).

The task is described as a reclassification because it involves revisiting the lexical category classification of roots found in \mapu{Mapudüngun} verb themes (see \S\ref{sec.02.1}) which was originally established by the anthropologist Ineke \citeauthor{smeets2008} in her work on the grammatical description of the \mapu{Mapuche} language, published in \citeyear{smeets2008} with the title ``\textit{A Grammar of Mapuche}''.

This research, like others conducted within the \mapu{Düngupeyüm}\footnote{\mapu{Düngupeyüm} is the name of the computational morphological analysis system that has been implemented; the name derives from \mapu{düngu-} `language, speech, concern', \mapu{-pe} suffix indicating physical or temporal proximity, \mapu{-yü} suffix indicating a habitual aspect feature, and the affix \mapu{-m}, a nominaliser indicating place or instrument. The translation would be something like `instrument always used for language', i.e., `language tool'. The system is publicly available on \url{https://www.chandia.net/dungupeyum/}\vspace{0.2cm}} project, relies on the grammatical description provided by \cite{smeets2008}. Our goal is to enhance the computational morphological analysis system that we have developed using finite-state transducers (\citealp{chandia2012}, \citeyear{chandia2021}, \citeyear{chandia2022}), which initially implemented the anthropologist's work. This focus on the elements and dynamics of the language allows us to refine and expand the capabilities of the system.

For a computational system such as the one described earlier, accurately determining both the lexical category of roots and the valency of verbal roots is crucial for proper analysis. This is because, in \mapu{Mapudüngun}, the addition of suffixes to verbal forms, particularly aspectual ones, depends on the verbal valency or its modification during word formation. Consequently, verbal valency plays a key role in analysing the form to identify the implicated suffixes. Therefore, it was necessary to verify the lexical category of roots involved in the formation of verbal themes, as \mapu{Mapudüngun} naturally verbalises roots from other lexical categories, including nouns, adjectives, adverbs, and even some pronouns and numerals. The aim is to accurately recognise the valency of roots that are genuinely verbal, avoiding misclassification of roots belonging to other categories. This document focuses on the task of lexical classification, while the classification of valency will be addressed in a subsequent study (\citealp{chandia2026}).

During the extensive period dedicated to developing the morphological analysis tool, several observations made by \cite{smeets2008} needed to be verified, and some required modification. She provided a grammatical description of \mapu{Mapudüngun} based on a group of four informants\footnote{As asserted by \citeauthor{smeets2008} in email correspondence, the number may be increased by four or five, which where sporadic informants, thus bringing the total to approximately ten.\vspace{0.2cm}}. As a result, her work does not fully capture the entirety of the \mapu{Mapuche} language or all the phenomena present within it.

The main contribution of \cite{smeets2008}' ``\textit{A Grammar of Mapuche}'', which is particularly valuable for the development of a computational morphological analysis system, is the identification of a prevalent or fixed position for approximately one hundred suffixes that can appear in \mapu{Mapuche} verbal forms. Additionally, she provides the morphotactics\footnote{\label{morfotactica}Morphotactics refers to the rules governing the allowed and prohibited combinations of morphemes (both among themselves and with roots) in a language. These rules influence certain forms within the language. For example, in a \mapu{Mapuche} nominal form, morphemes indicating verbal mode cannot appear unless the nominal form has been verbalised. Similarly, in a verbal form, a nominaliser affix and a grammatical person morpheme cannot co-occur. Morphotactic rules can be influenced by grammatical, semantic and morphological factors.\vspace{0.2cm}}, or combinatorial rules, for most of these suffixes.

In \mapu{Mapudüngun}, a minimal finite verbal form consists of a root (which can be verbal, nominal, adjectival, or adverbial), followed by one or more optional derivational suffixes, and at least one inflectional suffix. For a form to be finite, it must include a mode marker in slot four and a morpheme (which may be null) indicating the person in slot three.

\cite{smeets2008} states that suffixes appear in relatively fixed positions along the verbal form. Based on their placement relative to other suffixes and their function, these verbal affixes are organised into thirty-six numbered slots, arranged from right to left. Slot one is at the far right end of the form, while slot thirty-six is next to the root on the left. Certain slots contain suffixes that are mutually exclusive for grammatical or semantic reasons.

Although it is not unusual to find up to seven or eight suffixes following the root (see \hyperref[e1]{E.1}), verbs typically contain between four and six suffixes in spontaneous speech.

\paragraph{\example{1} \label{e1} 10 suffixes \mapu{Mapuche} verb \mapu{nünieñmarputueyiñmu}}~
\begin{enumerate}[label=\alph*.]
\vspace{-10pt}
\item[] \mapuex{nü \hspace{26pt} -nie \hspace{4pt} -ñma \hspace{2pt} -r \hspace{10pt} -pu \hspace{6pt} -tu \hspace{4pt} -e \hspace{10pt} -y \hspace{10pt} -ø \hspace{4pt} -iñ \hspace{4pt} -mu} \hfill (\citealp{smeets2008}: 443; 76)\\
\gloss{TV.take +PRPS +IO +ITR +LOC +RE +INV +IND +1 +PL +3A}\\
`they continued to take it away from us'; \textit{Literal:} `there they went continuously taking it away from us'
\end{enumerate}

\vspace{1cm}
\begin{figure}[htbp]
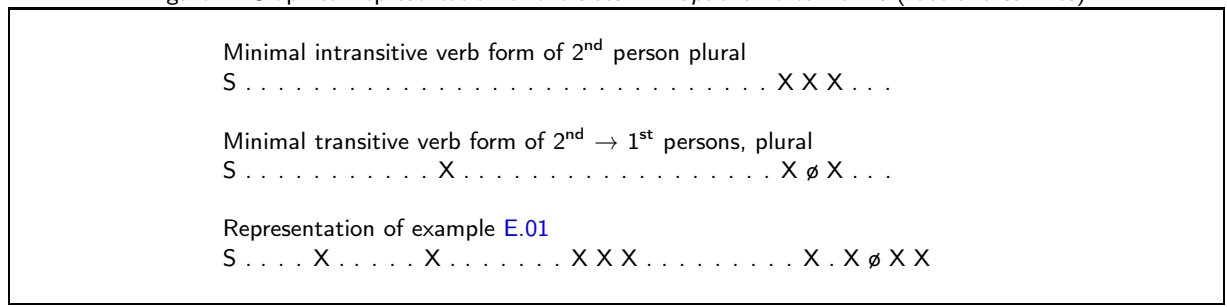

\centering
\begin{tabularx}{0.9\linewidth}{|X|}
\hline
\vspace{0.12cm} 
\qquad\qquad\qquad Minimal intransitive verb form of 2\textsuperscript{nd} person plural \\
\qquad\qquad\qquad S . . . . . . . . . . . . . . . . . . . . . . . . . . . . . . X X X . . .  \\\\

\qquad\qquad\qquad Minimal transitive verb form of 2\textsuperscript{nd} → 1\textsuperscript{st} persons, plural\\
\qquad\qquad\qquad S . . . . . . . . . . . X . . . . . . . . . . . . . . . . . . X ø X . . .  \\\\

\qquad\qquad\qquad Representation of \hyperref[e1]{E.1}\\
\qquad\qquad\qquad S . . . . X . . . . . X . . . . . . . X X X . . . . . . . . . X . X ø X X  \\\\
\hline
\end{tabularx}
\caption{\footnotesize Graphical representation of the slots in \mapu{Mapuche} verbal forms (root and suffixes)}
\label{fig:slots}
\end{figure}

\vspace{0.8cm}
In Figure~\ref{fig:slots}, three \mapu{Mapuche} verbal forms are graphically represented. In this representation, \textbf{S} denotes the verbal theme (stem), and each \textbf{dot} represents a slot. An \textbf{X} indicates a suffix realised in one of the slots, while \textbf{ø} also represents a suffix, specifically a null morpheme. Although the null morpheme has no phonemic or written form, it occupies a corresponding position or slot (\citealp{chandia2021}: 3).

In the \mapu{Mapuche} language, a verbal theme\footnote{To simplify, the term ``verbal theme'' refers to any root, whether simple or complex, to which suffixes are added to create a complete verbal predication. A simple ``verbal theme'' consists of a single root, while a complex ``verbal theme'' may involve two roots, a root with some suffixes, or a combination of these elements. Although it could be argued that these are more accurately described as lemmas, in this work they are all referred to as ``verbal theme'' for the sake of simplicity (\citealp{chandia2021}: 3).\vspace{0.2cm}} can consist of one, two, or occasionally three roots from different lexical categories, which may combine with each other and sometimes with certain suffixes. A verbal theme can be simple, containing just one root from any lexical category (\hyperref[e2a]{E.2a}, \hyperref[e2b]{E.2b}, \hyperref[e2c]{E.2c}), or complex, with several possible configurations:

\begin{itemize}
    \item A root combined with one or more suffixes (\hyperref[e2d]{E.2d}), sometimes up to three suffixes.
    \item A compound of two roots.
    \item A less common compound of three roots.
    \item A compound of two roots with suffixes, typically one but sometimes two or, exceptionally, three suffixes.
\end{itemize}
    
Suffixes can follow the first or second root in a compound, or even different suffixes follow each root. Additionally, verbal themes can include a reduplicated root, a root with a reduplicated suffix, or a reduplicated root followed by a suffix.

\paragraph{\example{2}}~
\vspace{-10pt}
\begin{enumerate}[label=\alph*.]
\item \label{e2a} \mapuex{pichi \hspace{22pt} -ka \hspace{14pt} -el} \hfill (\citealp{smeets2008}: 76; 16)\\
      \gloss{AJ.little +CONT +OVN}\\
      `[I] still was little (child)'
\item \label{e2b} \mapuex{düngu \hspace{20pt} -fi \hspace{8pt} -nge !} \hfill (\citealp{smeets2008}: 65; 33)\\
      \gloss{NN.speach +3P +IMP2SG}\\
      `talk to him/her'
\item \label{e2c} \mapuex{tripa \hspace{18pt} -y \hspace{10pt} -ø} \hfill (\citealp{smeets2008}: 67; 51)\\
      \gloss{IV.leave +IND +3}\\
      `he went out / he left'
\item \label{e2d} \mapuex{küpa \hspace{16pt} -l \hspace{12pt} -ün} \hfill (\citealp{deaugusta1903}: 295)\\
      \gloss{IV.come +CA +IND1SG}\\
      `I brought'; \textit{Literal:} `I made come'
\end{enumerate}

\subsection{\label{sec.01.1} Word Types In \mapu{Mapudüngun}: A State-Of-The-Art Review}
The classification of words into lexical types, or the use of a reliable existing classification, is essential in any computational approach to language involving analysis. This is particularly true in our case, which focuses on the morphological analysis of an agglutinative language, where roots of different Parts-of-Speech can generate an inflected verb form through the attachment of affixes.

\subsubsection{\label{sec.01.1.1} Dictionaries}
To date, the most reliable collections of words classified by their lexical type are the dictionaries published throughout the relatively short written history of the \mapu{Mapudüngun} language, particularly the following three\footnote{All three dictionaries are accessible on the CORLEXIM site: {\label{corlexim} \url{https://www.chandia.net/corlexim}} (\citealp{chandia2014}).\vspace{0.2cm}}:

\begin{enumerate}
    \item de Valdivia's dictionary, which reflects a vocabulary less influenced by Spanish, due to its early publication in 1606. A drawback for contemporary studies, however, is that a significant number of words in this dictionary are no longer in use or have undergone varying degrees of semantic and, to some extent, phonological change.
    \item Febrés' dictionaries, published in 1765 and 1846 respectively, reflect the influence of Spanish on the \mapu{Mapudüngun} language, resulting from over 300 years of contact between the \mapu{Mapuche} and Spanish cultures. These works often reveal fascinating shifts in meaning; for instance, some terms show notable semantic changes between the first and second dictionaries, or in comparison to Valdivia's earlier work. In some cases, these changes are intriguing, but at times, they seem disconcerting, as the shifts appear to stem from authorial errors. Nonetheless, Febrés' dictionaries remain valuable resources for contrast and confirmation in cases of uncertainty.
    \item de Augusta's dictionaries, published in 1916, are the most recent in time and offer a reflection of modern \mapu{Mapudüngun}. They clearly show the increased Spanish lexical influence after nearly 500 years of coexistence between the two cultures. This extended period of contact resulted in the imposition of Spanish culture over the \mapu{Mapuche} in almost every aspect of daily life, leading to a process of acculturation that negatively impacted both their culture and their language. The \mapu{Mapudüngun} language, in particular, was prohibited in the Chilean educational system. Despite these challenges, Augusta's dictionaries remain invaluable due to the remarkable linguistic rigour with which they were compiled.
    \item Havestadt's dictionary\footnote{Resource available for consultation via the following link: \url{https://benmolineaux.github.io/bookshelf/1777HAVE-Vocab.html}. An alternative Spanish version is also available at the following URL: \url{https://benmolineaux.github.io/bookshelf/1777Havestadt-ParsQuarta-MAPU-SPA.html} (\citealp{martinez-carmona2023}).\vspace{0.2cm}} is a fourth lexical source occasionally consulted in cases of doubt. Published in 1777, it provides definitions in Latin and can be a valuable resource, particularly when dealing with variations in forms.
    \end{enumerate}

Although these dictionaries are useful, the strong religious bias they carry must be acknowledged. They were compiled by figures such as the Spanish Jesuit de Valdivia, the Catalan Jesuit Febrés, the German Capuchin de Augusta (August Stephan Kathan), and the German Jesuit Havestadt. These works were primarily intended as tools for translating the Gospels and disseminating the Catholic faith, as well as for training priests who would carry out these tasks in \mapu{Mapuche} lands. A second bias stems from the Eurocentric view of the language, typical of earlier studies. This perspective prevailed until relatively recently when typological linguistic studies began to include new points of view that account for the unique features of languages which may differ significantly from the classical grammar models of the Western world.

As mentioned earlier, these dictionaries serve as a reliable source of lexically categorised words. In this regard, the Eurocentric perspective poses only a minor issue for our project, as we employ a broad classification of roots that generally aligns with Universal Grammar models\footnote{By ``Universal Grammar models'' we refer here to the working assumption that all languages share a basic set of lexical categories (noun, verb, adjective, adverb, etc.), not to any specific theoretical framework within generative grammar. This broad classification suffices for the computational purposes of this study.\vspace{0.2cm}}. The system does not utilise the specificity of idiosyncratic subtypes which seems unnecessary for the morphological analysis at this stage; however, they could prove useful for a more fine-grained grammatical analysis in the future.

However, not everything is straightforward. A significant limitation of the \mapu{Mapuche} dictionaries mentioned thus far lies in the way many of the entries are compiled. Often, they are collected with attached suffixes, meaning that the entry --and its lexical categorisation, when provided-- corresponds to the derived (and sometimes inflected) form rather than the isolated root. This practice complicates the task of determining the original category of the root, as its meaning --and occasionally its part of speech (PoS)-- can shift depending on the suffixes attached. This issue is further addressed in \S\ref{sec.02.1}.

\subsubsection{\label{sec.01.1.2} Computational Approach}
It is expected that any linguistic computational project considers the classification of the linguistic assets involved. This ensures a robust output from a linguistic perspective. While the system's functionality may not strictly require such classifications, when dealing with language, the goal is not only for the processes to function correctly but also for the results to be meaningful and useful at a linguistic level.

With this idea in mind, we explored the materials left by one of the few computational projects involving the \mapu{Mapuche} language, conducted in the early 2000s by a prestigious group of researchers from the Language Technologies Institute at Carnegie Mellon University and the Institute of Indigenous Studies at Universidad de La Frontera, supported by the Intercultural Bilingual Education Program of the Chilean Ministry of Education.

AVENUE was the name of this initiative, which produced a document entitled \textit{``Informe Final''} \citep{avenue2005} summarising the project's achievements and the works still to be done. When it started in 1999, it was known as NICE (Native-Language Interpretation and Communication Environment), later Avenue. Its goals were to create a simultaneous translation system and to contribute to the preservation and development of Indo-American languages\footnote{Avenue worked with three indigenous languages, setting up teams in three countries: Aymara in Bolivia, Quechua in Peru and \mapu{Mapudüngun} in Chile.\vspace{0.2cm}}. They were committed to creating their own alphabet for the language, transcribing texts according to this alphabet, extracting grammatical rules from the texts, compiling a parallel corpus of \mapu{Mapudüngun}-Spanish, also a glossary, developing a spell-checker of \mapu{Mapudüngun} for OpenOffice, and building a prototype of an automatic translator of \mapu{Mapudüngun}-Spanish. The report mentioned above states that these tasks were carried out at different levels of completion. The section dealing with the automatic translator mentions that it is an example-based machine translation system\footnote{Example-based machine translation (EBMT) works by searching its database for full translations or, more commonly, parts of sentences that have previously been translated by human translators.\vspace{0.2cm}}. At the time of writing, however, researchers say they are working on a transfer-based machine translation system\footnote{In a Transfer-Based Machine Translation (TBMT) system, the original text is first analysed morphologically and syntactically, resulting in a shallow syntactic representation. This representation is then transformed into a more abstract one that emphasises aspects relevant to the translation process and ignores other information. The transfer process converts this latter representation, still linked to the source language, into a representation at the same level of abstraction but linked to the target language. These two representations are called normalised or intermediate representations. From here the process is reversed: the syntactic components generate a representation of the text and finally the translation is generated.\vspace{0.2cm}} in which the basic constructions of \mapu{Mapudüngun} have been recorded: simple sentences with intransitive and transitive verbs, noun phrases with determiners and modifiers, verb phrases with different temporal and aspectual specifications, passive verb phrases, and verb phrases with inverse agreement\footnote{The inversion or reverse form in \mapu{Mapudüngun} is one of the possible expressions of the transitive verb (the others are direct and passive constructions) in which there is a third person acting upon a first or second person, this is expressed by the suffix \mapu{-e} before the mood suffix, and the suffix \mapu{-(m)ew} at the end, provided that the first or second person is a patient; i.e. the order of the constituents, which in \mapu{Mapudüngun} are suffixes, is not changed, but the detailed suffixes are added and combined to change the direction of the action of the verb (\citealp{zuniga2006}: 114).\vspace{0.2cm}}. Also relevant is the section dealing with the morphological analyser, which, according to the information in the report, would be an FST system fed by 105 suffixes and some 700 verbs. The document does not specify whether these ``verbs'' are only verb roots or also other type of roots, such as nominals or adjectivals, nor how the parser works in detail. According to some information that has been gathered, the parser would be developed using the Perl programming language and would also be able to generate sentences in \mapu{Mapudüngun}.

Among the extensive materials, there are indeed lists of lexemes, primarily derived from the dictionaries mentioned in the previous section \S\ref{sec.01.1.1}. Many of these lists adopt the classifications provided in those sources, while others reflect classifications informed by the linguistic knowledge of \mapu{Mapuche} participants and informants involved in the project. However, the project does not clarify the criteria used to establish the parts of speech (PoS) of the listed roots. Consequently, when comparing these lists with the dictionaries, it is impossible to determine which classification is more accurate.

\subsubsection{\label{sec.01.1.3} A Lexical Study}
In her thesis work \cite{villena2016} states that her research focuses on examining the authenticity and efficacy of the processes employed in \mapu{Mapudüngun} noun formation. Her objective is to develop criteria for the creation of new nouns in this language. In order to accomplish this objective, she has conducted an analysis of the same dictionaries referenced in \S\ref{sec.01.1.1}.

From the initial statement, it could be inferred that an accurate classification of roots would be of use and performed as a fundamental task. However, the abstract fails to mention that the noun formation under investigation is mainly by derivation processes (compounding is also taken into account), that is to say, by the mechanisms this language commonly uses for the purpose of identifying entities\footnote{From our point of view, these are not strictly neologisms, but available (morpho-syntactic) strategies that this language has to deal with the naming of things, new or known. For example, it would not be considered an English neologism: `Internet surfer', even though `Internet' is already a neologism, but `Internaut' would be; it is not registered but possible. What in English or Spanish would require a syntactic phrase —e.g., `the one who surfs the internet' / \textit{`el que navega por internet'}— in \mapu{Mapudüngun} can be expressed as a single derived word through regular suffixation. Calling such forms `neologisms' imposes an analytic-language perspective onto a polysynthetic one, where word-internal morphology does the work of syntactic constructions.\vspace{0.2cm}}. In any case, the subject of neologicity filters is addressed further down in the work, with the assertion that one of the criteria for determining whether a form is neological is the internal structure of the form (ibid: 11). This suggests that the root component of the neologism should be accurately lexically classified.

However, the prospect of a ``roots classification process'' appears to have been definitively lost after this note: ``In order to confirm the appropriateness of the lexical categories assigned by Augusta, the most adequate procedure would have been to verify their functioning in discourse. However, the size of the corpus to be analysed made this impossible'' (ibid: note 82, p. 114). --In the following point \S\ref{sec.02.1} we explain the procedure followed in our work, which actually implies the process mentioned in \cite{villena2016}'s note--.

Moreover, it appears that accurate lexical classification is not of paramount importance in this research study, particularly in the context of composition, as evidenced by the following citation: ``This finding led to the conclusion that the categories of roots involved in compounding are not crucial in determining the structure of \mapu{Mapudüngun} compounds. Instead, the type of semantic condition maintained between them is deemed to be more significant''\footnote{As \cite{villena2016} notes, this is a conclusion of Baker and Fasola (2009) that is supported by the observation of a similarity in the order of the V-N and N-N compounds.\vspace{0.2cm}} (ibid: 63).

Despite the aforementioned observations, further explorations were conducted on this work, with the objective of identifying any roots that might have undergone a verification of their lexical type. However, ultimately, we were only able to find confirmation of a task that had not been realised, as evidenced by the following few examples:

\paragraph{\example{3} \label{e3}} \hfill {\scriptsize(\citealp{villena2016})}
\vspace{-10pt}\begin{enumerate}[label=\alph*.]
    \item \label{e3a} \mapuex{\textbf{küdaw}-fe} \hspace{20pt} (\mapu{küdaw-} `to work') \hspace{24pt}  `worker'
    \item \label{e3b} \mapuex{\textbf{pu}-pülli-amuy} \hspace{4pt} (`gone into the ground') \hspace{4pt} `a certain healing art of \mapu{machi}\footnote{A \mapu{machi} is the \mapu{Mapuche} chaman.\vspace{0.2cm}}'
    \item \label{e3c} \mapuex{\textbf{yewen}} \hspace{30pt} (`shame') \hspace{58pt} `male or female genitalia'
\end{enumerate}

Example \hyperref[e3a]{E.3a} ((6.b), p. 53 in the source) shows \mapu{küdaw-} as a verbal root being nominalised by the suffix \mapu{-fe}, which the author says that combines with verbal bases to generate nouns that denote a characteristic agent. But in the dictionaries consulted on the CORLEXIM website (\citealp{chandia2014}) the following entries are found:

\begin{enumerate}
\item \mapu{\textbf{küdaw}} s. The work. \hfill ({\scriptsize de Augusta map-cas in \citealp{chandia2014}})
\item \mapu{\textbf{küdaw}} N. work. \hfill ({\scriptsize Smeets map-ing in \citealp{chandia2014}})
\item \mapu{\textbf{küdaw}} work. \hfill ({\scriptsize Febrés map-cas in \citealp{chandia2014}})
\item \textbf{(to be) engrossed} adj. \mapu{ngoymaduamkülen}; \hfill ({\scriptsize de Augusta cas-map in \citealp{chandia2014}})\\
\mapu{ngoymakonkülen kiñe \textbf{küdaw} mew} `to be engrossed in a task". 
\end{enumerate}

As demonstrated above, three authors identify the form as a noun; de Augusta provides an example of the use of the word `\textit{absorto}' in \mapu{Mapudüngun} (being engrossed in a job), where \mapu{küdaw} is used as a noun without any suffixes, which is in contrast to the classification used by \cite{villena2016}. This example is one of many such instances that can be found in her study.

\hyperref[e3b]{E.3b} ((92.g), p. 170 in the source) has been identified as an example of sintagmation, and it is said to represent the Prep-N-V structure. However, a review of dictionaries reveals that:

\begin{quote}
\mapu{\textbf{pu-}} Vi. to arrive; also \mapu{puw-}. The distribution of \mapu{pu(w)-} is not entirely clear. \mapu{puw-} is most frequent. I found \mapu{pu-fu-n} `I arrived, but' (RR) and \mapu{iñche pu-n} `I arrived' (JM), and I found \mapu{puw-küle-} (RR) only once; \mapu{müna ayuwüy iñchiu yu puwel} `he was very glad that we had arrived'; \mapu{küpa pulen liwen} `I want to arrive early'; \mapu{petu puwlaymi küdawmew} `you are too young to work' (Smeets map-ing in \citealp{chandia2014}.)
\end{quote}

Literally translated, \hyperref[e3b]{E.3b} would mean something like: `arrived at the ground as he was going' or `reached the ground on his way'. Perhaps there has been some confusion with another homophonic particle which is a prepositive and which de Augusta and Febrés, among others, define as follows:

\begin{quote}
\mapu{\textbf{pu}} artic. pref. pluralises the names of people, animals and some things. \mapu{pu wentru} `the men'. \mapu{pu  wangülen} `the stars' (de Augusta map-cas in \citealp{chandia2014}); see \citealp{deaugusta1903}: 15, 1.

\mapu{\textbf{pu}} (part. bef. nouns) denotes plurality; item, at or in; by; under; e.g. \mapu{pu ofida}, sheep (pl.); \mapu{pu ruka}, at home, in the house; \mapu{pu liwen}, in the morning; \mapu{pu minue}, under the ground (Febrés map-cas in \citealp{chandia2014}).
\end{quote}

In addition, there is a high probability that the translocative suffix \mapu{-pu} derives from the same verb, \mapu{pu-} mentioned above. The affix indicates that the denoted event takes place at a location away from the speaker. This function has also been described by Febrés and others:

\begin{quote}
\mapu{\textbf{pu, rpu}} (interp. to verbs) is to carry out the action of the verb by going from here to there. , v. gr. \mapu{pepuafin} or \mapu{perpuafin}, I'll drop by to see him; \mapu{pipuafin} or \mapu{pirpuafin}, I'll pass on to say to him (Febrés map-cas in \citealp{chandia2014})
\end{quote}

\hyperref[e3c]{E.3c} ((98.b), p. 176 in the source) is employed to illustrate a case of metonymy, yet it transpires to be a case of semantic calque from the Spanish `las vergüenzas'. In point of fact, the online DLE dictionary provides the eighth definition for the entry ``vergüenza'' as follows:

\begin{quote}
f. pl. Órganos sexuales externos del ser humano. La pintura muestra a Adán con una hoja de parra cubriendo sus vergüenzas. (REAL ACADEMIA ESPAÑOLA: Diccionario de la lengua española, 23.ª ed., [versión 23.8 en línea]. <https://dle.rae.es> [2024-12-21].)
\end{quote}

In conclusion, it is evident that the extant literature provides lists of the lexical classification of \mapu{Mapudüngun} roots that are neither complete nor totally reliable. While lexicons represent the most valuable resource in this respect, the remaining works also offer valuable insights that can facilitate the resolution of ambiguities in numerous cases.

\subsection{\label{sec.01.2} Summary of the Work}

This study is a systematic reclassification of the lexical categories of roots involved in \mapu{Mapudüngun} verbal themes. The primary goal of the work is to establish a reliable foundation for valency classification (\citealp{chandia2026}) by ensuring that roots identified as verbal are indeed inherently verbal, rather than verbalised forms belonging to other lexical categories (e.g., nouns, adjectives, adverbs).

The methodology employed is a straightforward verification process: roots listed as verbal in \cite{smeets2008}'s dictionary are cross-referenced against their usage in isolation in historical dictionaries (de Augusta, Febrés, de Valdivia) and the autobiography of the \mapu{longko Paskwal Koña} (\citealp{mosbach1930}). If a root is found used without derivational suffixes in these sources, it is inferred that its original category is non-verbal. This process yielded a comprehensive table (\hyperref[tab3]{Table 3} in the \nameref{sec.04}) showing 238 roots initially classified as verbs but reclassified as nouns, adjectives, adverbs, or demonstratives.

Key findings from this study include:
\begin{itemize}
    \item A significant number of roots in \cite{smeets2008}'s verbal inventory are verbalised forms of other lexical categories.
    \item Lists of verbal roots that co-occur with the causative suffix \mapu{-(ü/e)l} and \mapu{-(ü)m} were compiled from the \mapu{longko Koña}'s memoirs (\citealp{mosbach1930}).
    \item A set of labile roots (those that can be used both transitively and intransitively without morphological change) was identified.
\end{itemize}

The following article (\citealp{chandia2026}) builds directly on this foundation. The lexical classifications established here are taken as the starting point for the analysis that follows.

\section{\label{sec.02} Verbal Roots}

As mentioned in the \nameref{sec.01} (\S\ref{sec.01}), for a computational system like the one developed within this project, correctly identifying the valency of verbal roots is essential for computational morphological analysis. Many of the rules governing the combination of roots and suffixes depend on this classification or on changes in value during the word formation process. This is particularly important for achieving an unambiguous analysis or minimising ambiguity\footnote{In a computational morphological analysis system, ambiguity is related to the number of analyses generated by the machine. In other words, the more possible analyses the analyser returns for a lexical form, the more ambiguous the form is. Typically, the sentence context resolves the ambiguity, but this task must be handled by another system: a grammatical disambiguator. However, this tool is only helpful when analysing sentences; for isolated words, or words without context, it is less useful.\vspace{0.2cm}}. As stated in the introduction, before undertaking the task of valency classification for \mapu{Mapudüngun} verbal roots, a (re)categorisation of the roots that form verbal themes in this language will be carried out. The aim is to more accurately determine the lexical category of these roots, which will help in identifying the roots to be included in the subsequent work on recognising verbal valency (\citealp{chandia2026}). From a computational perspective, this will also enable a better morphological analysis of \mapu{Mapuche} verbs.

\subsection{\label{sec.02.1} Lexical Category of Roots}

The verbal roots listed by \cite{smeets2008} in her dictionary were examined to verify their lexical classification and to modify it if they actually belong to other lexical categories and have been verbalised instead. To achieve this, we primarily referred to the dictionaries by de Augusta, Febrés, and de Valdivia through the CORLEXIM (\citealp{chandia2014}) platform, as well as to the autobiography of the \mapu{longko}\footnote{Literally, \mapu{longko} means `head', but similar to other languages, such as Catalan, it is used to denote a chief or leader. In the \mapu{Mapuche} context, a \mapu{longko} is a leader or chief in a broad sense, essentially a community guide who ensures the well-being of his \mapu{lof} (community). On the other hand, a \mapu{toki} is the chief in matters of war. In a more everyday sense, \mapu{toki} means `axe'.\vspace{0.2cm}} \mapu{Paskwal Koña}\footnote{\label{pkoña}The \mapu{longko Paskwal Koña} narrated his life to the Capuchin missionary Alois Wilhelm von Mösbach (padre Ernesto), who transcribed it in \mapu{Mapudüngun}. This took place in the 1920s when \mapu{Koña} was still an energetic \mapu{longko} of 80 years. This autobiography is a unique document, not only for the information it provides about the \mapu{Mapuche} language and customs, but also because it is one of the few written sources in \mapu{Mapudüngun} containing words dictated by a \mapu{Mapuche} speaker.\vspace{0.2cm}} (\citealp{mosbach1930}). Examples of the roots to be verified were sought in all these sources. If any root was found used in isolation —i.e., not as part of a verbal form, nor with affixes attached— it was inferred that the original form is nominal, adjectival, or adverbial, depending on the context and usage. Many of these roots also appear in \cite{smeets2008}'s dictionary with the category listed in the column `Revised Cat.' in \hyperref[tab1]{Table 1}. These are non-verbal lexical categories, indicating that these words have been verbalised. In other words, the author records the meaning of the form in its original category as well as in its verbalised version. This column also reflects the lexical category reached after verifying each root in this study. The categories of the roots extracted from \cite{smeets2008}'s dictionary are listed in the `Initial Cat.' column. Some of these roots are recorded by \cite{smeets2008} with an incorporated suffix, which can alter their meaning and valency when the suffix is absent. These suffixes are extracted from the original form and listed in the `Extracted suffixes' column —the following \hyperref[tab1]{Table 1} is only a sample available for easy verification. The complete table, which contains two hundred and thirty-eight roots, can be found in the \nameref{sec.04}, \S\ref{sec.04}, \hyperref[tab3]{Table 3}.

\vspace{0.8cm}
\begin{longtable}{@{}p{2.2cm}p{1.6cm}p{1.3cm}p{1.4cm}p{7.3cm}@{}}
\caption{Verbal Roots Reclassified as Other Parts of Speech (Sample)}
\label{tab1}\\
\toprule
\textbf{Root} & \textbf{Extracted Suffixes} & \textbf{Original Cat.} & \textbf{Revised Cat.} & \textbf{Definition}\\
& & \scriptsize(Smeets) & \scriptsize(this study) & \\
\midrule
\endfirsthead
\multicolumn{5}{c}{\tablename\ \thetable\ -- \textit{Continued}}\\
\toprule
\textbf{Root} & \textbf{Extracted Suffixes} & \textbf{Original Cat.} & \textbf{Revised Cat.} & \textbf{Definition}\\
& & \scriptsize(Smeets) & \scriptsize(this study) & \\
\midrule
\endhead
\bottomrule
\endfoot
\bottomrule
\endlastfoot
\mapu{ad-} & & \textsc{IV} & \textsc{N} & the visible or characteristic aspects of something or someone, or of a group\\
\mapu{chadi-} & & \textsc{IV} & \textsc{N} & salt\\
\mapu{dakel-} & & \textsc{TV} & \textsc{N} & commission, agreement\\
\mapu{echid- $\sim$ echiw-} & & \textsc{IV} & \textsc{N} & sneeze\\
\mapu{fa-} & & \textsc{IV} & \textsc{Dem} & this, here\\
\mapu{ina-} & & \textsc{IV} & \textsc{Av} & close\\
\mapu{kachu-} & & \textsc{IV} & \textsc{N} & grass, herb\\
\mapu{la-} & & \textsc{IV} & \textsc{Aj} & dead, deceased\\
\mapu{llag-} & & \textsc{TV} & \textsc{Aj} & half\\
\mapu{machi-} & & \textsc{IV} & \textsc{N} & shaman\\
\mapu{nag-} & & \textsc{IV} & \textsc{Av} & down, below, under, beneath\\
\mapu{ngeñi-} & \mapu{-ka} & \textsc{IV} & \textsc{Aj} & urgent, pressing\\
\mapu{ñochi-} & & \textsc{IV} & \textsc{Aj} & peaceful, calm\\
\mapu{paf-} & & \textsc{IV} & \textsc{N} & boil, abscess\\
\mapu{raküm-} & & \textsc{TV} & \textsc{N} & closure, wall, protection\\
\mapu{tonon-} & & \textsc{TV} & \textsc{N} & weft\\
\mapu{traf-} & & \textsc{IV} & \textsc{Aj} & together, gathered, joined, fitted, interlocked\\
\mapu{uma- $\sim$ umañ- $\sim$ umag- $\sim$ umaw-} & & \textsc{IV} & \textsc{N} & sleepiness, drowsiness\\
\mapu{üna-} & & \textsc{IV} & \textsc{N} & itch\\
\mapu{wachi-} & & \textsc{TV} & \textsc{N} & trap\\
\mapu{yafü-} & & \textsc{IV} & \textsc{Aj} & hard, firm\\
\end{longtable}
\begin{center}
\begin{minipage}{14cm}
{\footnotesize \textbf{Abbreviations:} \textsc{IV} = Intransitive Verb; \textsc{TV} = Transitive Verb; \textsc{N} = Noun; \textsc{Aj} = Adjective; \textsc{Av} = Adverb;\\ \textsc{Dem} = Demonstrative.}
\end{minipage}
\end{center}

\subsection{\label{sec.02.2} Valency of Verbal Roots}

The second step in the classification of roots that participate in \mapu{Mapuche} verbal themes is to specify the valency of verbal roots; as previously mentioned, this task will be presented in \cite{chandia2026}. However, we will preview the results of this study and present a list of some verbal roots that co-occur with the causative suffix \mapu{-(ü/e)l} (see \S\ref{sec.02.3}). We also provide a list of labile roots (see \S\ref{sec.02.4}) and a table of roots that undergo causativisation with the causative suffix \mapu{-(ü)m} (\hyperref[tab4]{Table 4}, for a quick sample refer to \hyperref[tab2]{Table 2}).

\subsection{\label{sec.02.3} Verbal Roots That Co-occur With The Causative Suffix \mapu{-(ü/e)l}}

In the autobiography of the \mapu{longko} (\citealp{mosbach1930}), twenty-one verbal roots\footnote{Approximately one hundred and ten verbal roots that were not forming compounds were checked. Since this task took too long, it was decided to postpone the verification of roots when they participate in verbal compounds and exhibit a suffix in one of the forms that the causative can take: \mapu{-l}, \mapu{-ül}, \mapu{-el}.\vspace{0.2cm}} were found that combine with the causative \mapu{-l}, as shown in the list below. Almost all the roots are intransitive, except for two that form both transitive and intransitive verbs, meaning that they are labile or bivalent roots (for a detailed explanation see the companion valency article, \citealp{chandia2026}). In the following list, the two senses of the labile verbs are marked according to their valency, with \textsc{IV} (intransitive verb) and \textsc{TV} (transitive verb).

\begin{enumerate}[label=\arabic*.]
\item \mapu{aku-} `to arrive'
\item \mapu{allfü-} `to get injured, hurt, damaged'
\item \mapu{amu-} `to go, to go on'
\item \mapu{apo-} `to get filled'
\item \mapu{kümtrü-} `to rumble, to resonate'
\item \mapu{küpa-} `to come'
\item \mapu{miaw-} `to prowl, to roam'
\item \mapu{monge-} \textsc{IV} `to live, to heal, to get healed' / \textsc{TV} `to revive, to heal'
\item \mapu{montu-} `to escape'
\item \mapu{nepe-}\footnote{\label{nepe}The computational system generates all possible analyses, which sometimes leads to conjectures about the origin of certain \mapu{Mapudüngun} forms. For instance, \mapu{nepe-} meaning `to wake up' might have originated from the compound \mapu{ne-} $\sim$ \mapu{nie-} + \mapu{pe}: `to have' + `sight, vision'. Similarly, it can be inferred that \mapu{pewma} meaning `dream' comes from \mapu{pe-} + \mapu{uma(g)}: `sight, vision' + `sleep'.\vspace{0.2cm}} `to wake up'
\item \mapu{pültrü-} `to hang'
\item \mapu{rapi-} `to vomit'
\item \mapu{reyü-} `to mix, to blend'
\item \mapu{ru-} `to go along, to go through, to pass'
\item \mapu{rümu-} `to sink, to plunge, to immerse oneself'
\item \mapu{rünga-} `to embed, to become encrusted'
\item \mapu{tripa-} `to leave, to go out, to depart, to set off'
\item \mapu{ürfi-} `to drown, to sink'
\item \mapu{wüda-} `to divide, to split up, to separate'
\item \mapu{wüño-} `to return, to start again'
\item \mapu{yewe-} \textsc{IV} `to feel embarrassed, to be ashamed' / \textsc{TV} `to respect'
\end{enumerate}

\subsection{\label{sec.02.4} Collecting Labile Roots In \mapu{Düngupeyüm}}

In the \mapu{Düngupeyüm} morphological analysis system, labile roots will now appear in both the transitive and intransitive lists, which inevitably leads to increased ambiguity when analysing verb forms; however, excluding these roots would omit potential analyses. Additionally, the system has been updated to ensure that causatives, whether \mapu{-(ü)m} or \mapu{-(ü)l}, do not co-occur with transitive roots. Thus, causatives will only combine with labile roots in their intransitive sense. The following list comprises the labile verb roots that have been incorporated into the system:

\paragraph*{List of labile roots collected in the system}
\vspace{-0.5em}
\begin{enumerate}[label=\arabic*.]
\item \mapu{aye-} `to laugh' (\textsc{IV})\\
      \mapu{aye-} `to laugh at; to laugh with' (\textsc{TV})
      \vspace{0.2em}
\item \mapu{kewa-} `to fight, to quarrel, to struggle, to combat' (\textsc{IV})\\
      \mapu{kewa-} `to hit, to strike, to defeat, to overcome' (\textsc{TV})
      \vspace{0.2em}
\item \mapu{llüka-} `to get frightened, to be scared, to fear, to be afraid' (\textsc{IV})\\
      \mapu{llüka-} `to scare, to frighten' (\textsc{TV})
      \vspace{0.2em}
\item \mapu{meke-} `to take care of, to deal with, to be busy, to last, to endure' (\textsc{IV})\\
      \mapu{meke-} `to make' (\textsc{TV})
      \vspace{0.2em}
\item \mapu{monge-} `to get life, to heal, to recover' (\textsc{IV})\\
      \mapu{monge-} `to revive, to bring back to life, to heal' (\textsc{TV})
      \vspace{0.2em}
\item \mapu{nge-} `to be' (\textsc{IV})\\
      \mapu{nge-} `to have' (\textsc{TV}) (see \S3.4.2.3 in \citealp{chandia2026})
      \vspace{0.2em}
\item \mapu{püna-} `to stick, to glue, to adhere' (\textsc{IV})\\
      \mapu{püna-} `to glue, to paste' (\textsc{TV})
      \vspace{0.2em}
\item \mapu{waychüf-} `to fall, to get turned over, to roll, to spin' (\textsc{IV})\\
      \mapu{waychüf-} `to discard, to flip, to roll, to spin' (\textsc{TV})
      \vspace{0.2em}
\item \mapu{yewe-} `to feel embarrassed, to be ashamed' (\textsc{IV})\\
      \mapu{yewe-} `to respect' (\textsc{TV})
\end{enumerate}

\subsection{\label{sec.02.5} Verbal Roots That Co-occur With The Causative Suffix \mapu{-(ü)m}}

In the companion valency article (\citealp{chandia2026}: \S3.3), it will be shown that verifying the presence of the causative suffix \mapu{-(ü)m} in \mapu{Mapuche} verb forms is a reliable method for determining that the root carrying it is intransitive. However, it is also necessary to verify whether the root is genuinely verbal or belongs to another lexical category, a task that this document addresses. Furthermore, as noted by \cite{longkon2011a}, \cite{golluscio2007} and \cite{smeets2008}, not all intransitive roots accept causativisation with \mapu{-(ü)m}. Therefore, it is important to examine these authors' claims regarding the role of telicity or atelicity in the selection of the causative suffix. In \hyperref[tab4]{Table 4} (\nameref{sec.04}), all the roots that have been verified to combine with the causative suffix \mapu{-(ü)m} are presented. The following \hyperref[tab2]{Table 2} is only a sample available for easy verification of what is stated in the valency article.

\begin{longtable}{@{}p{1.6cm}p{0.8cm}p{3.6cm}p{1.8cm}p{3.4cm}ccc@{}}
\caption{Roots That Causativise With \mapu{-(ü)m} (Sample)}
\label{tab2}\\
\toprule
\textbf{Root} & \textbf{Cat.} & \textbf{Intransitive sense} & \textbf{Root+CA} & \textbf{Transitive sense} & \multicolumn{3}{c}{\textbf{Collected from}\textsuperscript{\hyperlink{tab2na}{\hypertarget{otab2na}{a}}}}\\
\midrule
\endfirsthead
\multicolumn{7}{c}{\tablename\ \thetable\ -- \textit{Continued}}\\
\toprule
\textbf{Root} & \textbf{Cat.} & \textbf{Intransitive sense} & \textbf{Root+CA} & \textbf{Transitive sense} & \multicolumn{3}{c}{\textbf{Collected from}}\\
\midrule
\endhead
\bottomrule
\endfoot
\bottomrule
\endlastfoot
\mapu{ad} & \textsc{N} & form, place, familiar, face, image, custom & \mapu{ad-üm} & to be experienced/skilled (in something) & K. & S. & \\
\mapu{anü} & \textsc{IV} & to get settled, to get sit & \mapu{anü-m} & to settle, to set, to put, to place & K. & S. & G.\\
\mapu{fa} & \textsc{Dem} & this, here & \mapu{fa-m} & to be like this & K. & S. & \\
\mapu{firkü} & \textsc{Aj} & fresh, cool & \mapu{firkü-m} & to cool, to warm slightly, to take the chill off & K. & & \\
\mapu{kon} & \textsc{IV} & to enter, to begin & \mapu{kon-üm} & to make come in, to add & K. & & G.\\
\mapu{la} & \textsc{Aj} & dead, deceased & \mapu{la\textbf{ng}-üm} & to kill & K. & S. & G.\\
\mapu{(l)el} & \textsc{IV} & to let oneself, to become loose, to detach & \mapu{(l)el-üm} & to release, to let go, to kill & K. & & \\
\mapu{llum} & \textsc{IV} & to get hidden, to conceal oneself & \mapu{llum-üm} & to hide, to conceal & K. & & \\
\mapu{nel} & \textsc{Aj} & loose, untied, unleashed & \mapu{nel-üm $\sim$ nel\textbf{k}-üm} & to release, to let go, to free & & S. & \\
\mapu{nor} & \textsc{Aj} & straight, correct, right & \mapu{nor-üm} & to correct, to rectify, to straighten & & S. & \\
\mapu{pewü} & \textsc{IV} & to swirl, to gather in a whirl, to twist, to sprain, to coil, to curl up & \mapu{pewü-m} & to measure, to reach out using the arms, to twist, to turn & K. & & \\
\mapu{pümü} & \textsc{Aj} & tense, tight, taut, strained & \mapu{pümü-m} & to stretch, to tighten, to tense & K. & & \\
\mapu{puw} & \textsc{IV} & to arrive, to reach, to achieve, to stay, to remain & \mapu{puw-üm} & to send, to deliver, to convey, to serve & K. & & \\
\mapu{trar} & \textsc{IV} & to ooze, to suppurate, to secrete & \mapu{trar-üm} & to drain the pus & & S. & \\
\mapu{tüng} & \textsc{IV} & to calm down, to take time & \mapu{tüng-üm} & to leave alone, not to disturb & K. & & \\
\mapu{üy} & \textsc{N} & name & \mapu{üy-üm} & to name, to denominate & & S. & \\
\mapu{wim} & \textsc{Aj} & accustomed, used to, meek, tame, domesticated & \mapu{wim-üm} & to make accustomed, to domesticate, to tame & & S. & \\
\mapu{yung} & \textsc{Aj} & pointed, sharp & \mapu{yung-üm} & to sharpen, to make pointed & K. & & \\
\end{longtable}
\begin{center}
\begin{minipage}{14cm}
{\footnotesize \hypertarget{tab2na}{\hyperlink{otab2na}{a}} The ``Collected from'' column indicates which source contains examples of the root with the causative suffix.\\ \textbf{K.} = \citealp{mosbach1930} (autobiography of \mapu{longko Paskwal Koña}); \textbf{S.} = \citealp{smeets2008} (A Grammar of \mapu{Mapuche}); \textbf{G.} = \citealp{golluscio2007} (Gramática descriptiva del \mapu{mapuche}).}
\end{minipage}
\end{center}

\section{\label{sec.03} Conclusions}

In the present work, we undertook a reclassification of the lexical categories of roots involved in verbal themes of the \mapu{Mapuche} language, which are part of the inventory of examples in the descriptive grammar of \mapu{Mapudüngun} developed by the anthropologist Ineke \cite{smeets2008}. This is a preliminary task to valency classification (\citealp{chandia2026}), the study that motivated us to begin this lexical category review. The task of valency classification of \mapu{Mapuche} verbal roots required an accurate classification of the lexical categories of the roots to avoid attributing verbal features, such as valency, to elements that cannot exhibit them because they are not verbal. A simple verification method for the lexical category was devised, as explained in \S\ref{sec.02.1}, and despite its simplicity, it has proven to be effective, with results shown in \hyperref[tab3]{Table 3} (\nameref{sec.04}).

Next, as the two studies referred to in this article are now complete, we present some of the overall material here, as we have included some of the data generated on completion of both studies in this initial document.

In a computational system like an automatic morphological analyser, where it is crucial to have specific and non-redundant data, ambiguity can be significantly reduced if roots are listed, as much as possible, under a single classification per form. Despite this, we chose to list roots with ambivalent behaviour (see \S\ref{sec.02.4}) in both lists, intransitive verbal roots and transitive verbal roots, with each list reflecting the root's valency-related sense. This approach might introduce varying degrees of ambiguity in the analyses produced by the machine. However, besides the ambiguity, it also allows for potential analyses that would be excluded if ambivalent roots were not collected in this manner.

It is essential to first identify the category most likely to be the source of the multiple interpretations or derivations that a form, used as the main component of the verbal theme, might generate (see \S\ref{sec.02.1}). Therefore, the primary task is to ensure that valency values are sought in inherently verbal roots rather than in verbalised roots derived from other categories (see \hyperref[tab3]{Table 3} in the \nameref{sec.04}). This can be challenging in a language like \mapu{Mapudüngun}, as there is often insufficient material or stored knowledge to determine the etymology of much of its vocabulary, largely due to historical and social factors. Techniques like the one employed in this study, while reflecting a highly conditioned perspective, often diverge from fully verifiable data. Nonetheless, it is at least known that \cite{mosbach1930} is one of the best references for the \mapu{Mapuche} language (see footnote \ref{pkoña}), which was the main reason for its use in this research.

To conclude, the decision to publish the study in two parts —lexical classification presented here and valency classification presented in \citealp{chandia2026}— aims to clarify the findings. As stated in the summary, ``The outcomes of this categorical and valency review directly benefit the computational analyser, as they are implemented as soon as they are verified. Additionally, these findings are expected to enhance the understanding of the \mapu{Mapuche} language''. Readers are invited to review the accompanying article to this first work, which provides a comprehensive understanding of the results of this research.

\newpage
\section*{Acknowledgements}

First and foremost, I express my gratitude for the always timely and appropriate support and advice from my thesis supervisors, Irene Castellón and Elizabeth Comelles. Without their guidance, the form and structure of both articles would be far from achieving a good presentation and understandable order, and the terminology used would not always have been accurate.

Secondly, my deepest thanks go to Fernando Zúñiga, who reviewed and corrected not only the content aspects of both articles but also some formal aspects. Without his extensive linguistic knowledge in general and of \mapu{Mapudüngun} in particular, many of the points discussed along the two articles would have been nothing more than imprecise approximations to the linguistic ideas and situations of the \mapu{Mapuche} language addressed in both works.

\vspace{2cm}
\section{\label{sec.04} Appendix}

\subsection{\label{sec.04.1} Labels And Abbreviations}

\begin{multicols}{2}
\setlength{\columnseprule}{0.1pt}
\noindent
\textsc{1, 2, 3} \hfill First, Second, Third persons, respectively\\
\textsc{1t2A} \hfill First agent to Second patient\\
\textsc{3A} \hfill Third person agent\\
\textsc{3P} \hfill 3\textsuperscript{rd} person patient (Differential Object Marker)\\
\textsc{ADJDO} \hfill Doable action adjectiviser\\
\textsc{Aj, AJ} \hfill Adjective\\
\textsc{AP} \hfill Anaphora\\
\textsc{Av, AV} \hfill Adverb\\
\textsc{BEN} \hfill Benefactive\\
\textsc{CA} \hfill Causative\\
\textsc{Conj, CJ} \hfill Conjunction\\
\textsc{CONT} \hfill Continuative\\
\textsc{CR.IV} \hfill Intransitive compound verbal stem\\
\textsc{CR.TV} \hfill Transitive compound verbal stem\\
\textsc{Dem, DP} \hfill Demonstrative\\
\textsc{DL} \hfill Dual number\\
\textsc{EXP} \hfill Experimentative\\
\textsc{FAC} \hfill Factual\\
\textsc{FORCE} \hfill Obligative\\
G. \hfill Lucía Golluscio\\
\textsc{FUT} \hfill Future\\
\textsc{HAB} \hfill Habituative\\
\textsc{IND} \hfill Indicative\\
\textsc{IND1SG} \hfill Indicative 1\textsuperscript{st} person singular\\
\textsc{INST} \hfill Instrumentaliser\\
\textsc{Int, IP} \hfill Interrogative\\
\textsc{INV} \hfill Transitive inversion marker\\
\textsc{IO} \hfill Indirect object\\
\textsc{ITR} \hfill Interruptive\\
\textsc{IVN} \hfill Instrumental verbal noun\\
K. \hfill \mapu{Paskwal Koña}\\
\textsc{LOC} \hfill Locative\\
\textsc{MIO} \hfill More implied object\\
\textsc{N, NN} \hfill Noun (substantive)\\
\textsc{NEG} \hfill Negator\\
\textsc{NOM} \hfill Nominaliser\\
\textsc{NU} \hfill Numeral\\
\textsc{OO} \hfill Oblique object\\
\textsc{OVN} \hfill Objective verbal noun\\
\textsc{PASS} \hfill Passive voice\\
\textsc{PFPS} \hfill Perfect persistent \\
\textsc{PL} \hfill Plural number\\
\textsc{PLR} \hfill Pluraliser\\
\textsc{PRPS} \hfill Progressive persistent\\
\textsc{PVN} \hfill Plain verbal noun\\
\textsc{RE} \hfill Repetitive/Restaurative\\
\textsc{REF} \hfill Reflexive/Reciprocal\\
\textsc{RI} \hfill Ruptured implicature\\
\textsc{RNNR} \hfill Reduplicated nominal root\\
S. \hfill Ineke Smeets\\
\textsc{SFR} \hfill Stem formant\\
\textsc{SG} \hfill Singular number\\
\textsc{SJI} \hfill Subjunctive affix in imperatives forms\\
\textsc{SP} \hfill Possessive pronoun\\
\textsc{ST} \hfill Estative\\
\textsc{SVN} \hfill Subjective verbal noun\\
\textsc{TH} \hfill Thither\\
\textsc{TR} \hfill Transitiviser\\
\textsc{IV} \hfill Intransitive verb\\
\textsc{VRB} \hfill Verbaliser\\
\textsc{TV} \hfill Transitive verb
\end{multicols}

\vspace{2cm}
\subsection{\label{sec.04.2} Tables}

\begin{longtable}{@{}p{2.2cm}p{1.6cm}p{1.3cm}p{1.4cm}p{7.3cm}@{}}
\caption{Verbal Roots Reclassified as Other Parts of Speech}
\label{tab3}\\
\toprule
\textbf{Root} & \textbf{Extracted Suffixes} & \textbf{Original Cat.} & \textbf{Revised Cat.} & \textbf{Definition}\\
&  & \scriptsize(Smeets) & \scriptsize(this study) & \\
\midrule
\endfirsthead
\multicolumn{5}{c}{\tablename\ \thetable\ -- \textit{Continued}}\\
\toprule
\textbf{Root} & \textbf{Extracted Suffixes} & \textbf{Original Cat.} & \textbf{Revised Cat.} & \textbf{Definition}\\
& & \scriptsize(Smeets) & \scriptsize(this study) & \\
\midrule
\endhead
\bottomrule
\endfoot
\bottomrule
\endlastfoot
\mapu{ad-} & & \textsc{IV} & \textsc{N} & the visible or characteristic aspects of something or someone, or of a group\\
\mapu{af-} & & \textsc{IV} & \textsc{N} & end\\
\mapu{afü-} & & \textsc{IV} & \textsc{Aj} & cooked, ripe\\
\mapu{allush-} & & \textsc{IV} & \textsc{Aj} & lukewarm\\
\mapu{alü-} & & \textsc{IV} & \textsc{Av} & very, much\\
\mapu{angim-} & & \textsc{TV} & \textsc{N} & jerky\\
\mapu{angka-} & & \textsc{IV} & \textsc{N} & middle, half\\
\mapu{antü-} & & \textsc{IV} & \textsc{N} & Sun, day, time\\
\mapu{are-} & & \textsc{IV} & \textsc{Aj} & hot, warm\\
\mapu{ariñ-} & & \textsc{IV} & \textsc{Aj} & burnt, smoked\\
\mapu{arof-} & & \textsc{IV} & \textsc{N} & sweat\\
\mapu{asul-} & & \textsc{IV} & \textsc{Aj} & blue\\
\mapu{awi-} & & \textsc{IV} & \textsc{Aj} & hot\\
\mapu{awka-} & & \textsc{IV} & \textsc{Aj} & rebellious, wild\\
\mapu{aylen-} & & \textsc{IV} & \textsc{N} & ember(s)\\
\mapu{aywiñ-} & & \textsc{IV} & \textsc{N} & shade\\
\mapu{chadi-} & & \textsc{IV} & \textsc{N} & salt\\
\mapu{chafo-} & & \textsc{IV} & \textsc{N} & cough\\
\mapu{chali-} & & \textsc{IV} & \textsc{N} & greeting\\
\mapu{challwa-} & & \textsc{IV} & \textsc{N} & fish\\
\mapu{chape-} & & \textsc{TV} & \textsc{N} & braid (in the hair)\\
\mapu{che-} & & \textsc{IV} & \textsc{N} & person, people\\
\mapu{chedkuy-} & & \textsc{TV} & \textsc{N} & father/son in law (of a man)\\
\mapu{chilla-} & & \textsc{TV} & \textsc{N} & saddle, chair\\
\mapu{chod-} & & \textsc{IV} & \textsc{Aj} & yellow\\
\mapu{chum-} & & \textsc{TV} & Int & how, what\\
\mapu{chüngküd-} & & \textsc{IV} & \textsc{Aj} & circular, rounded\\
\mapu{dakel-} & & \textsc{TV} & \textsc{N} & commission, agreement\\
\mapu{dew-} & & \textsc{IV} & \textsc{Av} & already, since, because, no more, only, finally\\
\mapu{domo-} & & \textsc{IV} & \textsc{N} & woman\\
\mapu{doy- $\sim$ yod-} & & \textsc{TV} & \textsc{Av} & more\\
\mapu{duam-} & & \textsc{TV} & \textsc{N} & matter, issue, need, intention\\
\mapu{dumdum-} & & \textsc{IV} & \textsc{N} & penumbra\\
\mapu{düngu-} & & \textsc{TV} & \textsc{N} & issue, matter, subject, concern, word, language\\
\mapu{echid- $\sim$ echiw-} & & \textsc{IV} & \textsc{N} & sneeze\\
\mapu{elfal-} & & \textsc{TV} & \textsc{N} & commission\\
\mapu{eñum-} & & \textsc{IV} & \textsc{Aj} & comfortingly/pleasantly warm\\
\mapu{epew-} & & \textsc{IV} & \textsc{N} & tale, story, history, narration\\
\mapu{fa-} & & \textsc{IV} & \textsc{Dem} & this, here\\
\mapu{fe-} & & \textsc{IV} & \textsc{Dem} & that, there\\
\mapu{fente-} & & \textsc{TV} & \textsc{Av} & so, much\\
\mapu{fentre-} & & \textsc{IV} & \textsc{Av} & very, much\\
\mapu{fey-} & & \textsc{IV} & \textsc{Dem} & this, that, she, he\\
\mapu{filla-} & & \textsc{IV} & \textsc{N} & scarcity, lack\\
\mapu{fitrun-} & & \textsc{IV} & \textsc{N} & smoke\\
\mapu{fücha-} & & \textsc{IV} & \textsc{Aj} & big, old\\
\mapu{fülang-} & & \textsc{IV} & \textsc{Aj} & white\\
\mapu{funa-} & & \textsc{IV} & \textsc{Aj} & rotten\\
\mapu{füre-} & & \textsc{IV} & \textsc{Aj} & spicy, hot\\
\mapu{futrül-} & & \textsc{TV} & \textsc{N} & heap, pile\\
\mapu{füw-} & & \textsc{TV} & \textsc{N} & thread, wool\\
\mapu{ina-} & & \textsc{IV} & \textsc{Av} & close\\
\mapu{ingka-} & & \textsc{TV} & \textsc{N} & defender\\
\mapu{kachu-} & & \textsc{IV} & \textsc{N} & grass, herb\\
\mapu{kafe-} & & \textsc{IV} & \textsc{N} & coffee\\
\mapu{kallfü-} & & \textsc{IV} & \textsc{Aj} & blue, purple, violet\\
\mapu{kangka-} & & \textsc{TV} & \textsc{Aj} & roasted\\
\mapu{karü-} & & \textsc{IV} & \textsc{Aj} & green\\
\mapu{kashü-} & & \textsc{IV} & \textsc{Aj} & gray\\
\mapu{katrü-} & & \textsc{TV} & \textsc{Aj} & cut off, interrupted\\
\mapu{kawe-} & & \textsc{IV} & \textsc{N} & oar\\
\mapu{kellu-} & & \textsc{TV} & \textsc{N} & help\\
\mapu{kelü-} & & \textsc{IV} & \textsc{Aj} & red\\
\mapu{kim-} & & \textsc{TV} & \textsc{Aj} & wise, knower\\
\mapu{kochü-} & & \textsc{IV} & \textsc{Aj} & sweet\\
\mapu{kofke-} & & \textsc{TV} & \textsc{N} & bread\\
\mapu{kolü-} & & \textsc{IV} & \textsc{Aj} & rust, reddish-brown (colour)\\
\mapu{kompañ-} & & \textsc{TV} & \textsc{N} & partner\\
\mapu{koñoll-} & & \textsc{IV} & \textsc{Aj} & purple, violet\\
\mapu{koñü-} & & \textsc{IV} & \textsc{N} & baby, newborn\\
\mapu{korü-} & & \textsc{IV} & \textsc{N} & soup, broth\\
\mapu{kotrü-} & & \textsc{IV} & \textsc{Aj} & salty, sour, bitter\\
\mapu{kowkow-} & & \textsc{IV} & \textsc{N} & owl\\
\mapu{küdaw-} & & \textsc{IV} & \textsc{N} & work\\
\mapu{kude-} & & \textsc{IV} & \textsc{N} & bet\\
\mapu{kufü- $\sim$ kufüll-} & & VI/TV & \textsc{Aj} & spoiled, damaged (due to the effect of heat)\\
\mapu{küllay-} & \mapu{-tu} & \textsc{TV} & \textsc{N} & quillay\textsuperscript{\hyperlink{tab3na}{\hypertarget{otab3na}{a}}}, quillay soap\\
\mapu{külü-} & & \textsc{IV} & \textsc{Aj} & tilted, slanted\\
\mapu{küme-} & & \textsc{IV} & \textsc{Aj} & good\\
\mapu{kunaw-} & & \textsc{IV} & \textsc{Aj} & inflated, floating\\
\mapu{kupül-} & & \textsc{TV} & \textsc{Aj} & fixed, secure\\
\mapu{kuram-}\textsuperscript{\hyperlink{tab3nb}{\hypertarget{otab3nb}{b}}} & & \textsc{IV} & \textsc{N} & egg\\
\mapu{küri-} & & \textsc{IV} & \textsc{Aj} & lustful, lecherous\\
\mapu{kurü-} & & \textsc{IV} & \textsc{Aj} & black\\
\mapu{kürüf-} & & \textsc{IV} & \textsc{N} & wind\\
\mapu{kushe-} & & \textsc{IV} & \textsc{Aj} & elderly, old (woman)\\
\mapu{kutran-} & & \textsc{IV} & \textsc{N} & illness, disease, pain, suffering\\
\mapu{kütrüng-} & & \textsc{TV} & \textsc{N} & bundle, bunch\\
\mapu{küyen-} & & \textsc{IV} & \textsc{N} & moon, month\\
\mapu{la-} & & \textsc{IV} & \textsc{Aj} & dead, deceased\\
\mapu{lantra-} & & \textsc{IV} & \textsc{Aj} & bulging, thick\\
\mapu{lawen-} & & \textsc{TV} & \textsc{N} & medicine (medicinal herb)\\
\mapu{lef-} & & \textsc{IV} & \textsc{Aj} & fast (going/moving)\\
\mapu{leg-} & & \textsc{IV} & \textsc{Aj} & certain, accurate, correct\\
\mapu{lichi-} & & \textsc{IV} & \textsc{N} & milk\\
\mapu{lif-} & & \textsc{IV} & \textsc{Aj} & clean\\
\mapu{ling-ar $\sim$ lüng-af $\sim$ lüg- $\sim$ lig-} & \mapu{-ar, -af}\textsuperscript{\hyperlink{tab3nc}{\hypertarget{otab3nc}{c}}} & \textsc{IV} & \textsc{Aj} & white, clear\\
\mapu{llag-} & & \textsc{TV} & \textsc{Aj} & half\\
\mapu{llekü-} & & \textsc{IV} & \textsc{Av} & close\\
\mapu{llitu-} & & \textsc{IV} & \textsc{Aj} & initial, first\\
\mapu{lliwa- $\sim$ lluwa-} & \mapu{-tu} & \textsc{TV} & \textsc{N} & soothsayer, seer\\
\mapu{llüngüd-} & & \textsc{TV} & \textsc{Aj} & flat, level\\
\mapu{llüpañ-} & & \textsc{TV} & \textsc{Aj} & lying, reclining\\
\mapu{lüf-} & & \textsc{IV} & \textsc{Aj} & scorched, roasted\\
\mapu{lüp-}\textsuperscript{\hyperlink{tab3nd}{\hypertarget{otab3nd}{d}}} & \mapu{-üm} & \textsc{TV} & \textsc{Aj} & scorched, roasted\\
\mapu{lüykü-} & & \textsc{IV} & \textsc{N} & drop\\
\mapu{machi-} & & \textsc{IV} & \textsc{N} & shaman\\
\mapu{madom- $\sim$ masom-} & \mapu{tuku-} & \textsc{TV} & \textsc{N} & ember, charred log\\
\mapu{mafü-} & & \textsc{TV} & \textsc{N} & dowry\\
\mapu{mallma- $\sim$ malma-} & & \textsc{IV} & \textsc{Aj} & boastful, arrogant, pretentious\\
\mapu{mangkü-} & & \textsc{TV} & \textsc{N} & kick, boot\\
\mapu{mañum-} & & \textsc{TV} & \textsc{N} & gratitude, thankfulness\\
\mapu{mawün-} & & \textsc{TV} & \textsc{N} & rain\\
\mapu{may-} & & \textsc{TV} & Conj & yes\\
\mapu{me-} & & \textsc{IV} & \textsc{N} & excrement, scum, dross\\
\mapu{melkay-} & & \textsc{IV} & \textsc{Aj} & slippery\\
\mapu{meñku- $\sim$ menku-} & & \textsc{TV} & \textsc{N} & load, burden\\
\mapu{minggako-} & & \textsc{IV} & \textsc{N} & communal/community work\\
\mapu{misha-} & & \textsc{TV} & \textsc{N} & dining companion\\
\mapu{mollfü- $\sim$ mollfüñ-} & & \textsc{IV} & \textsc{N} & blood\\
\mapu{motri-} & & \textsc{IV} & \textsc{Aj} & fat\\
\mapu{moy-} & & \textsc{IV} & \textsc{N} & boil\\
\mapu{moyo-} & & \textsc{TV} & \textsc{N} & breast\\
\mapu{muday-} & & \textsc{IV} & \textsc{N} & chicha\\
\mapu{mukur- $\sim$ mukud-} & & \textsc{IV} & \textsc{Aj} & sour, bitter\\
\mapu{müna-} & \mapu{-le} & \textsc{IV} & \textsc{Av} & many, much\\
\mapu{mupi-} & & \textsc{IV} & \textsc{Aj} & true, genuine\\
\mapu{müpü-} & & \textsc{IV} & \textsc{N} & wing\\
\mapu{mürke- $\sim$ murke-} & & \textsc{TV} & \textsc{N} & toasted flour\\
\mapu{mütrüm-} & & \textsc{TV} & \textsc{N} & call\\
\mapu{nag-} & & \textsc{IV} & \textsc{Av} & down, below, under, beneath\\
\mapu{nel- $\sim$ ney-} & & \textsc{IV} & \textsc{Aj} & loose, unfastened\\
\mapu{ngeñi-} & \mapu{-ka} & \textsc{IV} & \textsc{Aj} & urgent, pressing\\
\mapu{ngoyma-} & & \textsc{TV} & \textsc{Aj} & unproductive, neglected\\
\mapu{ngüdi-} & & \textsc{TV} & \textsc{N} & obstruction\\
\mapu{ngülam-} & & \textsc{TV} & \textsc{N} & advice, tip\\
\mapu{ngüñü-} & & \textsc{IV} & \textsc{N} & hunger\\
\mapu{nor-} & & \textsc{IV} & \textsc{Aj} & straight (in a physical or figurative sense), correct \\
\mapu{notu-} & \mapu{-ka} & \textsc{TV} & \textsc{N} & obstinacy, blindness\\
\mapu{nüla-} & & \textsc{TV} & \textsc{Aj} & open\\
\mapu{nümü-} & & \textsc{IV} & \textsc{N} & smell, odour\\
\mapu{nüngay- $\sim$ ningay-} & & \textsc{IV} & \textsc{N} & anger, fury, rage, impatience\\
\mapu{nürüf-} & & \textsc{TV} & \textsc{Aj} & closed\\
\mapu{nütram-} & & \textsc{TV} & \textsc{N} & recount, conversation, story\\
\mapu{nüyün-} & & \textsc{IV} & \textsc{N} & tremor, quake\\
\mapu{ñochi-} & & \textsc{IV} & \textsc{Aj} & peaceful, calm\\
\mapu{ñum-} & & \textsc{IV} & \textsc{N} & upper millstone\\
\mapu{paf-} & & \textsc{IV} & \textsc{N} & boil, abscess\\
\mapu{pakar- $\sim$ pakaka-} & & \textsc{IV} & \textsc{N} & a large toad\\
\mapu{pali-} & & \textsc{IV} & \textsc{N} & palin\textsuperscript{\hyperlink{tab3ne}{\hypertarget{otab3ne}{e}}} ball\\
\mapu{panü-} & & \textsc{TV} & \textsc{N} & fathom\\
\mapu{papel-} & & \textsc{TV} & \textsc{N} & paper\\
\mapu{payun- $\sim$ payuin- $\sim$ payum- $\sim$ payuñ- $\sim$ payun'} & & \textsc{IV} & \textsc{N} & beard\\
\mapu{perimo-} & \mapu{-ntu} & \textsc{IV} & \textsc{N} & omen\\
\mapu{perkü-} & & \textsc{IV} & \textsc{N} & fart\\
\mapu{pewma-} & & \textsc{IV} & \textsc{N} & daydream, dream\\
\mapu{pichi- $\sim$ püchi- $\sim$ püchü- $\sim$ püti- $\sim$ pütü-} & & \textsc{IV} & \textsc{Aj} & little\\
\mapu{pidku-} & & \textsc{TV} & \textsc{Aj} & stew (referring to a dish typically involving legumes cooked in a pot)\\
\mapu{piku-} & & \textsc{IV} & \textsc{N} & North\\
\mapu{pilu-} & & \textsc{IV} & \textsc{Aj} & deaf\\
\mapu{pire-} & & \textsc{IV} & \textsc{N} & snow, hail\\
\mapu{pitra- $\sim$ pitraw-} & & \textsc{IV} & \textsc{N} & blister, callus\\
\mapu{pitru-} & & \textsc{IV} & \textsc{N} & scabies\\
\mapu{pod-} & & \textsc{IV} & \textsc{Aj} & dirty\\
\mapu{pofre-} & & \textsc{IV} & \textsc{Aj} & poor\\
\mapu{poy-} & & \textsc{IV} & \textsc{N} & tumour, swelling\\
\mapu{püd-} & & \textsc{IV} & \textsc{Aj} & thick, dense, condensed\\
\mapu{pülle-} & & \textsc{IV} & \textsc{Av} & close, next\\
\mapu{püntü-} & & \textsc{IV} & \textsc{N} & piece, chunk\\
\mapu{püñeñ-} & & \textsc{IV} & \textsc{N} & daughter or son (to the mother or her sister)\\
\mapu{püro-} & & \textsc{TV} & \textsc{N} & knot\\
\mapu{pütre-} & & \textsc{TV} & \textsc{Aj} & blazing, scorching, burning, hot\\
\mapu{pütrü-} & & \textsc{IV} & \textsc{Av} & in abundance, a lot, plentifully\\
\mapu{raküm-} & & \textsc{TV} & \textsc{N} & closure, wall, protection\\
\mapu{rangiñ-} & & \textsc{IV} & \textsc{N} & middle, half\\
\mapu{re-} & \mapu{-le} & \textsc{IV} & \textsc{Aj} & alone, exclusive, pure\\
\mapu{riku-} & & \textsc{IV} & \textsc{Aj} & rich (it may be related to \mapu{rükü-} further down)\\
\mapu{ruka-} & & \textsc{TV} & \textsc{N} & house\\
\mapu{rükü-} & & \textsc{IV} & \textsc{Aj} & mean, petty, greedy, miserly, tight-fisted, stingy\\
\mapu{rümpel-} & & \textsc{IV} & \textsc{Aj} & jealous\\
\mapu{runa- $\sim$ truna-} & & \textsc{TV} & \textsc{N} & two-handed handful\\
\mapu{rutra-} & \mapu{-tu} & \textsc{TV} & \textsc{N} & pinch\\
\mapu{tonon-} & & \textsc{TV} & \textsc{N} & weft\\
\mapu{traf-} & & \textsc{IV} & \textsc{Aj} & together, gathered, joined, fitted, interlocked\\
\mapu{traf-iya-}\textsuperscript{\hyperlink{tab3nf}{\hypertarget{otab3nf}{f}}} & & \textsc{IV} & \textsc{Av} & last night, since last night's fall\\
\mapu{tralka-} & & \textsc{IV} & \textsc{N} & thunder, shotgun\\
\mapu{tran-} & & \textsc{IV} & \textsc{Aj} & fallen (especially referring to trees)\\
\mapu{trangliñ-} & & \textsc{IV} & \textsc{N} & frost\\
\mapu{trapel-} & & \textsc{TV} & \textsc{N} & cord, string\\
\mapu{trar-} & & \textsc{IV} & \textsc{N} & pus\\
\mapu{trari- $\sim$ trarü-} & & \textsc{TV} & \textsc{N} & ribbon, tape, belt\\
\mapu{traytray-} & & \textsc{IV} & \textsc{N} & gargle, or similar gargling sound\\
\mapu{trem-} & & \textsc{IV} & \textsc{Aj} & grown, adult, mature\\
\mapu{tremo-} & & \textsc{TV} & \textsc{Aj} & healthy, robust, splendid\\
\mapu{trintri-} & & \textsc{IV} & \textsc{Aj} & curly\\
\mapu{tripantu-} & & \textsc{IV} & \textsc{N} & year\\
\mapu{tromü-} & & \textsc{IV} & \textsc{N} & cloud\\
\mapu{tror-} & & \textsc{IV} & \textsc{N} & foam\\
\mapu{trüfon-} & & \textsc{IV} & \textsc{N} & cough\\
\mapu{trufür-} & & \textsc{IV} & \textsc{N} & dust\\
\mapu{trüg- $\sim$ trow- $\sim$ trig-} & & \textsc{IV} & \textsc{N} & crack\\
\mapu{trülke-} & & \textsc{TV} & \textsc{N} & leather\\
\mapu{trümfül- $\sim$ trünfül- $\sim$ trüfül-} & & \textsc{IV} & \textsc{Aj} & twisted, bent\\
\mapu{trür- $\sim$ mür- $\sim$ ür-} & & \textsc{IV} & \textsc{N} & pair, couple\\
\mapu{trutruka-} & & \textsc{IV} & \textsc{N} & a wind instrument like a long cornet\\
\mapu{tute-} & & \textsc{TV} & \textsc{Aj} & correct, accurate\\
\mapu{üküm-} & & \textsc{IV} & \textsc{N} & silence\\
\mapu{ül-} & & \textsc{IV} & \textsc{N} & song, singing\\
\mapu{üllüf- $\sim$ illüf-} & & \textsc{IV} & \textsc{Aj} & unfortunate, miserable, eventful\\
\mapu{uma- $\sim$ umañ- $\sim$ umag- $\sim$ umaw-} & & \textsc{IV} & \textsc{N} & sleepiness, drowsiness\\
\mapu{üna-} & & \textsc{IV} & \textsc{N} & itch\\
\mapu{üng-} & \mapu{-üm} & \textsc{TV} & \textsc{Av} & while\\
\mapu{ütrüf- $\sim$ utruf- $\sim$ ütüf- $\sim$ itrüf-} & & \textsc{TV} & \textsc{N} & throw (in a game)\\
\mapu{üy- $\sim$ güy-} & & \textsc{IV} & \textsc{N} & name\\
\mapu{üyag-} & \mapu{-tu} & \textsc{TV} & \textsc{Aj} & both\\
\mapu{wachi-} & & \textsc{TV} & \textsc{N} & trap\\
\mapu{waw-} & & \textsc{IV} & \textsc{N} & valley, lowland, low-lying area\\
\mapu{wechod-} & & \textsc{IV} & \textsc{N} & hole\\
\mapu{wedwed-} & & \textsc{IV} & \textsc{Aj} & mischievous\\
\mapu{welu-} & & \textsc{IV} & \textsc{Aj} & alternative, alternating\\
\mapu{wentru-} & & \textsc{IV} & \textsc{N} & man\\
\mapu{wew-} & & \textsc{TV} & \textsc{N} & victory, triumph, gain, profit\\
\mapu{wichü- $\sim$ wüchü- $\sim$ wüchür-} & & \textsc{TV} & \textsc{Aj} & twisted, crooked\\
\mapu{widi- $\sim$ widü- $\sim$ wüdü-} & & \textsc{TV} & \textsc{N} & clay vessel/container, pottery\\
\mapu{wif-} & & \textsc{IV} & \textsc{Aj} & wide, broad\\
\mapu{wikef- $\sim$ wikür- $\sim$ wünfü- $\sim$ wingür-} & & \textsc{IV/TV} & \textsc{Aj} & torn, broken\\
\mapu{wilüf-} & & \textsc{IV} & \textsc{Aj} & shiny\\
\mapu{wim-} & & \textsc{IV} & \textsc{Aj} & tame, accustomed\\
\mapu{wima-} & & \textsc{TV} & \textsc{N} & rod, staff\\
\mapu{witral-} & & \textsc{TV} & \textsc{N} & warp, loom\\
\mapu{wün-} & & \textsc{IV} & \textsc{N} & mouth, dawn, sunrise\\
\mapu{wüne-} & & \textsc{IV} & \textsc{Aj} & fore, anterior, superior\\
\mapu{wütre-} & & \textsc{IV} & \textsc{Aj} & cold\\
\mapu{yafü-} & & \textsc{IV} & \textsc{Aj} & hard, firm\\
\mapu{yall-} & & \textsc{TV} & \textsc{N} & son\\
\mapu{yiwül- $\sim$ yiwüll-} & & \textsc{TV} & \textsc{N} & shuttle (referring to the loom)\\
\mapu{yochi-} & & \textsc{IV} & \textsc{Aj} & tight, fitted\\
\mapu{yung-} & \mapu{-üm} & \textsc{TV} & \textsc{Aj} & pointed, sharp\\
\end{longtable}
\begin{center}
\begin{minipage}{14cm}
{\footnotesize \noindent
\hypertarget{tab3na}{\hyperlink{otab3na}{a}} \mapu{Küllay} is a tree endemic to the central region of Chile; scientific name \textit{Quillaja saponaria}, also known as `soapbark tree'.\\
\hypertarget{tab3nb}{\hyperlink{otab3nb}{b}} In conjugated (verbalised) forms, the root \mapu{kuram} `egg' is alternatively analysed as \textsc{-nn.küra\_stone+ca.m34...} by \mapu{Düngupeyüm}, that is to say, as \mapu{kura-m}, a form that may be interpreted as `caused to be a stone', which may well be the origin of \mapu{kuram} `egg'.\\
\hypertarget{tab3nc}{\hyperlink{otab3nc}{c}} In this case (\mapu{ling-ar-, lüng-af-}), the suffixes were not extracted but are simply presented here. They appear to be suffixes (\mapu{-ar-, -af-}) that imply a semantic change (direction) when added to the base form. They might be unproductive suffixes that have become lexicalised with these roots, or they could be mere phonological fillers. No references to these elements were found in the various studies on \mapu{Mapudüngun}. However, it should be noted that a radical change occurs in the root \mapu{lüg- $\sim$ lig-}, similar in nature to that explained in the companion valency article (\citealp{chandia2026}, E.06), though this is triggered by a suffix about which we have no information.\\
\hypertarget{tab3nd}{\hyperlink{otab3nd}{d}} The causative suffix \mapu{-(ü)m} triggers a radical change in some roots, causing those ending in \mapu{f} to shift to \mapu{p} (see \citealp{chandia2026}, E.06), indicating that this root (\mapu{lüp-}) is the same as \mapu{lüf-} above.\\
\hypertarget{tab3ne}{\hyperlink{otab3ne}{e}} \mapu{Palin} is a sport similar to field hockey, referred to as \textit{chueca} by Spaniards and Creoles.\\
\hypertarget{tab3nf}{\hyperlink{otab3nf}{f}} \mapu{trafiya} $\sim$ \mapu{trafüya} is a lexicalised compound form, likely derived from \mapu{traf-} `to come together' and \mapu{wiya} $\sim$ \mapu{wüya} `yesterday', meaning `the coming together of yesterday [with today]', referring to the period from the previous evening to the following morning, `last night'.\\
}
\end{minipage}
\end{center}

\newpage

\begin{longtable}{@{}p{1.6cm}p{0.8cm}p{3.6cm}p{1.8cm}p{3.4cm}ccc@{}}
\caption{Roots That Causativise With \mapu{-(ü)m}}
\label{tab4}\\
\toprule
\textbf{Root} & \textbf{Cat.} & \textbf{Intransitive sense} & \textbf{Root+CA} & \textbf{Transitive sense} & \multicolumn{3}{c}{\textbf{Collected from}\textsuperscript{\hyperlink{tab4na}{\hypertarget{otab4na}{a}}}}\\
\midrule
\endfirsthead
\multicolumn{7}{c}{\tablename\ \thetable\ -- \textit{Continued}}\\
\toprule
\textbf{Root} & \textbf{Cat.} & \textbf{Intransitive sense} & \textbf{Root+CA} & \textbf{Transitive sense} & \multicolumn{3}{c}{\textbf{Collected from}}\\
\midrule
\endhead
\bottomrule
\endfoot
\bottomrule
\endlastfoot
1. \mapu{ad}& \textsc{N} & form, place, familiar, face, image, custom & \mapu{ad-üm} & to be experienced/skilled (in something) & K. & S. & \\
2. \mapu{afü}& \textsc{Aj} & cooked, ripe & \mapu{afü-m} & to cook & K. & S. & G.\\
3. \mapu{ali}& \textsc{IV} & to get heated up & \mapu{ali-m} & to heat up & & S. & \\
4. \mapu{angkü}& \textsc{IV} & to get dry (up), to be sterile, to be arid & \mapu{angkü-m} & to dry (up) & K. & S. & \\
5. \mapu{anü}& \textsc{IV} & to get settled, to get sit & \mapu{anü-m} & to settle, to set, to put, to place & K. & S. & G.\\
6. \mapu{are}& \textsc{Aj} & hot, warm & \mapu{are-m} & to warm up & & S. & G.\\
7. \mapu{chaw}& \textsc{N} & father, man & \mapu{chaw-üm}\textsuperscript{\hyperlink{tab4nb}{\hypertarget{otab4nb}{b}}} & to incubate & K. & S. & \\
8. \mapu{che}& \textsc{N} & person, people, human being & \mapu{che-m} & to relate\textsuperscript{\hyperlink{tab4nc}{\hypertarget{otab4nc}{c}}} & K. & & \\
9. \mapu{chong}& \textsc{IV} & to go out, to extinguish, to dry up & \mapu{chong-üm} & to turn off, to extinguish, to dry & & S. & \\
10. \mapu{chu} & Int & interrogative particle & \mapu{chu-m} & to ask a question\textsuperscript{\hyperlink{tab4nd}{\hypertarget{otab4nd}{d}}} & K. & S. & \\
11. \mapu{fa}& \textsc{Dem} & this, here & \mapu{fa-m} & to be like this & K. & S. & \\
12. \mapu{fay}& \textsc{IV} & to ferment & \mapu{fay-üm} & to [make] ferment & & S. & \\
13. \mapu{fe}& \textsc{Dem} & that, there & \mapu{fe-m} & to be like that & K. & S. & \\
14. \mapu{fey}& \textsc{IV} & to get adapted, to fit & \mapu{fey-üm} & to adapt, to fit, to insert & & S. & \\
15. \mapu{firkü}& \textsc{Aj} & fresh, cool & \mapu{firkü-m} & to cool, to warm slightly, to take the chill off & K. & & \\
16. \mapu{fül}& \textsc{Av} & close & \mapu{fül-üm} & to bring closer, to approximate, to bring near & K. & S. & \\
17. \mapu{füy}& \textsc{IV} & to squeeze (itself), to hold on (to itself) & \mapu{füy-üm} & to tighten, to hold & K. & & \\
18. \mapu{illku}& \textsc{IV} & to get angry & \mapu{illku-m} & to contradict, to make angry & K. & & \\
19. \mapu{kofi}& \textsc{IV} & to get heat up, to get overheat & \mapu{kofi-m} & to heat, to overheat & & S. & \\
20. \mapu{kon}& \textsc{IV} & to enter, to begin & \mapu{kon-üm} & to make come in, to add & K. & & G.\\
21. \mapu{kudu}& \textsc{IV} & to lie (down), to go to bed & \mapu{kudu-m} & to lay out & K. & & \\
22. \mapu{küme}& \textsc{Aj} & good, nice, pleasant & \mapu{küme-m} & to add value, to improve & K. & & \\
23. \mapu{kümpu}& \textsc{N} & piece, chunk, bit & \mapu{kümpu-m} & to chop, to cut into pieces, to tear apart, to shred & K. & & \\
24. \mapu{küpa}& \textsc{IV} & to come & \mapu{küpa-m} & to bring, to introduce & K. & & \\
25. \mapu{la}& \textsc{Aj} & dead, deceased & \mapu{la\textbf{ng}-üm} & to kill & K. & S. & G.\\
26. \mapu{laf}& \textsc{Aj} & flat, level & \mapu{la\textbf{p}-üm} & to stretch, to extend & & S. & \\
27. \mapu{law}& \textsc{IV} & to lose hair/feathers & \mapu{law-üm} & to shave, to shear, to pluck & & S. & \\
28. \mapu{lef}& \textsc{Aj} & fast (on going or moving) & \mapu{le\textbf{p}-üm} & to make run (animals), to make flee, to make turn away & & S. & G.\\
29. \mapu{(l)el}& \textsc{IV} & to let oneself, to become loose, to detach & \mapu{(l)el-üm} & to release, to let go, to kill & K. & & \\
30. \mapu{lif}& \textsc{Aj} & clean & \mapu{li\textbf{p}-üm} & to clean & K. & & \\
31. \mapu{llangkü}& \textsc{IV} & to fall (down) & \mapu{llangkü-m} & to drop, to lose & K. & S. & \\
32. \mapu{lleg}& \textsc{IV} & to grow (up), to come up & \mapu{lle\textbf{k}-üm} & to raise, to cultivate, to grow (plants), to bring out & & S. & G.\\
33. \mapu{lliw $\sim$ llüw $\sim$ lluw}& \textsc{IV} & to get merged, to get melted, to get liquefied & \mapu{lliw-üm} & to merge, to melt, to liquefy & & & G.\\
34. \mapu{llum}& \textsc{IV} & to get hidden, to conceal oneself & \mapu{llum-üm} & to hide, to conceal & K. & & \\
35. \mapu{lüf}& \textsc{Aj} & burned, roasted & \mapu{lü\textbf{p}-üm} & to burn, to set fire, to roast & & S. & G.\\
36. \mapu{lüg}& \textsc{Aj} & white & \mapu{lüg-üm} & to whiten, to bleach, to lighten (in color) & & S. & \\
37. \mapu{lulu}& \textsc{N} & noise, crash, upheaval & \mapu{lulu-m} & to make noise, to cause crashing & K. & & \\
38. \mapu{nag}& \textsc{Av} & down, underneath & \mapu{na\textbf{k}-üm} & to go down, to lower, to descend & K. & S. & G.\\
39. \mapu{nel}& \textsc{Aj} & loose, untied, unleashed & \mapu{nel-üm $\sim$ nel\textbf{k}-üm} & to release, to let go, to free & & S. & \\
40. \mapu{neng}& \textsc{IV} & to move (oneself) & \mapu{neng-üm} & to move, to shake, to shudder, to direct & K. & & \\
41. \mapu{neykü}\textsuperscript{\hyperlink{tab4ne}{\hypertarget{otab4ne}{e}}}& \textsc{IV} & to unleash, to get free & \mapu{neykü-m} & to release, to let go, to free & & S. & \\
42. \mapu{ngül}& \textsc{IV} & to get gathered, to come together, to join & \mapu{ngül-üm} & to gather, to bring together, to collect, to harvest & K. & S. & \\
43. \mapu{nong}& \textsc{IV} & to get broken down, to get damaged, to get scorched & \mapu{nong-üm} & to spoil, to damage, to scorch & K. & & \\
44. \mapu{nor}& \textsc{Aj} & straight, correct, right & \mapu{nor-üm} & to correct, to rectify, to straighten & & S. & \\
45. \mapu{ñam}& \textsc{IV} & to get lost, to get astray, to get loose & \mapu{ñam-üm} & to lose, to release, to let go, to make disappear & & S. & G.\\
46. \mapu{ñif}& \textsc{IV} & to dry out & \mapu{ñi\textbf{p}-üm} & to dry & & S. & \\
47. \mapu{ñom}& \textsc{Aj} & meek, gentle, tame, docile, obedient & \mapu{ñom-üm} & to tame, to domesticate, to tame & K. & S. & \\
48. \mapu{pelo}\textsuperscript{\hyperlink{tab4nf}{\hypertarget{otab4nf}{f}}}& \textsc{N} & light, clarity & \mapu{pelo-m} & to illuminate, to light up, to turn on the light & & S. & \\
49. \mapu{pewü}& \textsc{IV} & to swirl, to gather in a whirl, to twist, to sprain, to coil, to curl up & \mapu{pewü-m} & to measure, to reach out using the arms, to twist, to turn & K. & & \\
50. \mapu{piwü}& \textsc{IV} & to dry out & \mapu{piwü-m} & to dry (up) & K. & S. & \\
51. \mapu{pod}& \textsc{N} & dirt, filth, uncleanliness & \mapu{pod-üm} & to dirty, to soil & & S. & \\
52. \mapu{poñü}& \textsc{N} & potato & \mapu{poñü-m} & feed with potatoes & K. & & \\
53. \mapu{püd}& \textsc{Aj} & thick, dense, scattered, dispersed & \mapu{püd-üm} & to spread, to scatter, to thicken, to extend, to spread out, to stretch, to ration, to distribute & K. & S. & \\
54. \mapu{pümü}& \textsc{Aj} & tense, tight, taut, strained & \mapu{pümü-m} & to stretch, to tighten, to tense & K. & & \\
55. \mapu{püna}& \textsc{IV} & to get stuck, to get glued, to get adhered & \mapu{püna-m} & to stick, to glue, to adhere & & S. & \\
56. \mapu{püra}& \textsc{IV} & to ascend, to go up, to climb & \mapu{püra-m}\textsuperscript{\hyperlink{tab4ng}{\hypertarget{otab4ng}{g}}} & to lift, to raise, to pick up, to keep in mind, to remember, to make evident, to highlight & K. & S. & G.\\
57. \mapu{püsha}& \textsc{IV} & to run out, to be finished & \mapu{püsha-m} & to finish, to end, to spend, to use up, to exhaust & & & G.\\
58. \mapu{pütre}& \textsc{Aj} & burning, red-hot & \mapu{pütre-m} & to smoke, to fumigate\textsuperscript{\hyperlink{tab4nh}{\hypertarget{otab4nh}{h}}} & K. & & \\
59. \mapu{puw}& \textsc{IV} & to arrive, to reach, to achieve, to stay, to remain & \mapu{puw-üm} & to send, to deliver, to convey, to serve & K. & & \\
60. \mapu{rüngü}& \textsc{IV} & to be pulverised, to be ground, to be crushed & \mapu{rüngü-m} & to grind, to mill, to crush & & S. & \\
61. \mapu{traf}& \textsc{Aj} & together, joined, gathered, fitted & \mapu{tra\textbf{p}-üm} & to gather, to collect, to put together, to grab, to catch, to pick up & K. & S. & \\
62. \mapu{tral}& \textsc{IV} & to get hit, to get bumped, to get dented, to get bruised, to collide, to crash, to clash & \mapu{tral-üm} & to hit, to strike, to beat & K. & & \\
63. \mapu{tran}& \textsc{Aj} & fallen (especially referring to trees) & \mapu{tran-üm} & to make fall, to knock down, to knock over, to topple, to lay down & & & G.\\
64. \mapu{trar}& \textsc{IV} & to ooze, to suppurate, to secrete & \mapu{trar-üm} & to drain the pus & & S. & \\
65. \mapu{trem}& \textsc{Aj} & grown, adult, mature & \mapu{trem-üm} & to raise, to breed & K. & S. & G.\\
66. \mapu{trof}& \textsc{IV} & to burst, to break, to crack, to snap & \mapu{tro\textbf{p}-üm} & to split, to break & & S. & \\
67. \mapu{trong}& \textsc{Aj} & thick, dense & \mapu{trong-üm} & to roof (houses)\textsuperscript{\hyperlink{tab4ni}{\hypertarget{otab4ni}{i}}} & K. & & \\
68. \mapu{trür}& \textsc{Aj} & even, level, tight, composed & \mapu{trür-üm} & to equalise, to compensate, to balance, to indemnify & K. & S. & \\
69. \mapu{tüng}& \textsc{IV} & to calm down, to take time & \mapu{tüng-üm} & to leave alone, not to disturb & K. & & \\
70. \mapu{üng}& \textsc{Av} & while & \mapu{üng-üm} & to wait & K. & & \\
71. \mapu{üre $\sim$ ürü}& \textsc{IV} & to get wet & \mapu{üre-m $\sim$ ürü-m} & to wet, to moisten, to soak & K. & S. & \\
72. \mapu{ürkü}& \textsc{IV} & to get tired & \mapu{ürkü-m} & to tire, to tire out & & & G.\\
73. \mapu{üy $\sim$ üyü}& \textsc{IV} & to catch fire, to ignite & \mapu{üy-üm $\sim$ üyü-m} & to set fire, to ignite & K. & S. & \\
74. \mapu{üy}& \textsc{N} & name & \mapu{üy-üm} & to name, to denominate & & S. & \\
75. \mapu{wadkü}& \textsc{IV} & to boil & \mapu{wadkü-m} & to boil, to bring to a boil & K. & S. & G.\\
76. \mapu{wallü}& \textsc{IV} & to get surrounded, to get encompassed, to get completed & \mapu{wallü-m} & to tell or recount in detail & K. & & \\
77. \mapu{wef}& \textsc{IV} & to appear, to show itself & \mapu{we\textbf{p}-üm} & to make appear, to find, to show & & & G.\\
78. \mapu{welli}& \textsc{IV} & to get empty & \mapu{welli-m} & to empty, to vacate & K. & S. & \\
79. \mapu{wim}& \textsc{Aj} & accustomed, used to, meek, tame, domesticated & \mapu{wim-üm} & to make accustomed, to domesticate, to tame & & S. & \\
80. \mapu{wüda} & \textsc{IV} & to divide, to split up & \mapu{wüda-m} & to divide, to distribute, to share, to split & K. & S. & \\
81. \mapu{wünü}& \textsc{IV} & to get stretched, to get elongated & \mapu{wünü-m} & to stretch, to lengthen & & S. & \\
82. \mapu{yall}& \textsc{N} & son (of a man) & \mapu{yall-üm} & to multiply, to engender & & S. & \\
83. \mapu{yewe}& \textsc{IV} & to feel embarrassed, to be ashamed & \mapu{yewe-m} & to embarrass & & & G.\\
84. \mapu{yung}& \textsc{Aj} & pointed, sharp & \mapu{yung-üm} & to sharpen, to make pointed & K. & & \\
\end{longtable}
\begin{center}
\begin{minipage}{14cm}
{\footnotesize  \noindent
\hypertarget{tab4na}{\hyperlink{otab4na}{a}} The ``Collected from'' column indicates which source contains examples of the root with the causative suffix.\\ \textbf{K.} = \citealp{mosbach1930} (autobiography of \mapu{longko Paskwal Koña}); \textbf{S.} = \citealp{smeets2008} (A Grammar of \mapu{Mapuche}); \textbf{G.} = \citealp{golluscio2007} (Gramática descriptiva del \mapu{mapuche}).\\
\hypertarget{tab4nb}{\hyperlink{otab4nb}{b}} The \mapu{Mapuche} use the term \mapu{chaw-üm} `fathering', or more literally `to cause to father', to refer to `incubating eggs'.\\
\hypertarget{tab4nc}{\hyperlink{otab4nc}{c}} In reality, to convey the meaning of `to relate', the compound \mapu{che-m-ye} needs to be formed, which is analysed as \textsc{-nn.che\_person\_people+ca.m+tv.ye\_to-carry}. This is the way to express a relationship with someone in \mapu{Mapudüngun}. For example, \mapu{laku-ye-fi-ñ} translates to `he is my paternal grandfather', which literally means `I carry him as a grandfather' (\citealp{becerra2011}: 115).\\
\hypertarget{tab4nd}{\hyperlink{otab4nd}{d}} In the case of interrogatives like ``what, which, how'', etc., starting with \mapu{chu-}, these are formed by adding other suffixes. For example, \mapu{chu-m-al} means `what for', \mapu{chu-m-nge-chi} means `how', \mapu{chu-m-ül} means `when', and so on.\\
\hypertarget{tab4ne}{\hyperlink{otab4ne}{e}} It is very likely that \mapu{neykü} is a lexicalised form, composed of the root \mapu{nel-} listed in line 39, and the unproductive suffix \mapu{-kü}, which is found in a few verbal roots but has not been documented or described in existing literature.\\
\hypertarget{tab4nf}{\hyperlink{otab4nf}{f}} \mapu{pelo} is a lexicalised nominalised verb, likely from \mapu{pe-} `sight, vision' + fossilised suffix \mapu{-lo} (no longer productive), meaning `light, clarity'. For the origin of similar lexicalised forms, see footnote \ref{nepe}. For an example of \mapu{pelo} in a verbal form, see the companion valency article (\citealp{chandia2026}, E.28a).\\
\hypertarget{tab4ng}{\hyperlink{otab4ng}{g}} The form \mapu{püra-m} generates compounds with many action verbs to indicate an upward direction, such as in \mapu{tralkatunpüramün}, which means `to shoot into the air'. It is also used figuratively, as in \mapu{kimpüramün}, meaning `to become aware'; literally: `to raise knowledge'.\\
\hypertarget{tab4nh}{\hyperlink{otab4nh}{h}} When \mapu{pütre} is nominalised as \mapu{pütrem} (\textsc{-aj.pütre\_blazing+ca.m+nom.ø}), it means `cigar' or `tobacco'.\\
\hypertarget{tab4ni}{\hyperlink{otab4ni}{i}} \mapu{trongum-} is to thatch a \mapu{ruka} (a traditional \mapu{Mapuche} house) with reed or tall grass; these materials must be placed very densely, ensuring a good level of thickness.\\
}
\end{minipage}
\end{center}

\newpage
\bibliographystyle{johd}
\bibliography{bib}
\end{document}